\documentclass[]{artificer}
\usepackage{microtype}
\usepackage{amsfonts}
\usepackage{graphicx}
\usepackage{booktabs}
\usepackage{wrapfig}
\usepackage{amsmath}
\usepackage{amsthm}
\usepackage{marvosym}
\usepackage{algorithm}

\usepackage{algorithmic}
\usepackage[capitalize,noabbrev]{cleveref}
\usepackage{enumitem}

\usepackage{tcolorbox}
\usepackage{xcolor}
\colorlet{tabcolor}{colorbg}
\newcommand{\hlfirst}[1]{\multicolumn{1}{@{}>{\columncolor{tabcolor}[0pt][\tabcolsep]}l}{#1}}
\newcommand{\hllast}[1]{\multicolumn{1}{>{\columncolor{tabcolor}[\tabcolsep][0pt]}c@{}}{#1}}

\newcommand{\ie}{\textit{i.e.}}

\newcommand{\method}{AlignGraft}

\newtheorem{proposition}{Proposition}

\title{Test-Time Weak-to-Strong Alignment: Transferring Implicit Rewards from Weak to Strong Flow Models}

\author[\spadesuit]{Xin Xie}
\author[\spadesuit]{Fan Zhang}
\author[{\textnormal{\Letter}}]{Dong Gong}
\affiliation[]{University of New South Wales (UNSW Sydney)}
\contribution{\textsuperscript{$\spadesuit$}Equal contribution.}
\contribution{\textsuperscript{\Letter}Corresponding author.}

\abstract{ 
Aligning a text-to-image generation flow model with a reward makes it follow objectives that the training data alone does not provide. Alignment fine-tuning delivers this by reinforcement learning (RL) or preference optimization, but it must be repeated for every checkpoint and returns a model fixed at the reward and strength it was trained with. Test-time alignment instead steers a frozen model during sampling, allowing task-specific and sample-specific guidance. Existing methods obtain this only by drawing the per-step signal from the reward function itself, through its gradient, or through a separately trained value function. We propose changing the supervision source: let a pair of weak models, not a reward function, supply the supervision. A source aligned model, kept together with its base as a source alignment pair, stores its training reward as an implicit, step-wise, KL-anchored signal expressed in the sampler's own coordinates. We explore whether this model-form supervision can cross scale, and show that it does: our method, \method{}, aligns a larger, frozen, never-tuned model by adding the pair's velocity difference during sampling. The transport is exact under a shared noising kernel and needs neither the reward nor its gradient at test time. The method has no schedules, only a single scalar that controls the alignment strength and can extrapolate it beyond that of the source alignment pair. Across image and video flow models (Stable Diffusion 3.5, FLUX, and Wan), the transfer lifts the frozen large model on preference, compositional, and text-rendering rewards, can exceed the source aligned model itself, and preserves the large model's fidelity at a small constant sampling overhead. Extensive experiments show that one alignment run on a weak model produces supervision that the whole model family can reuse at test time.
}

\metadata[Project Page]{\href{https://artificer-ai-lab.github.io/AlignGraft/}{Artificer-AI-Lab/AlignGraft}}

\begin{document}

\maketitle

\begin{figure}[h]
\centering
\includegraphics[width=1.0\textwidth]{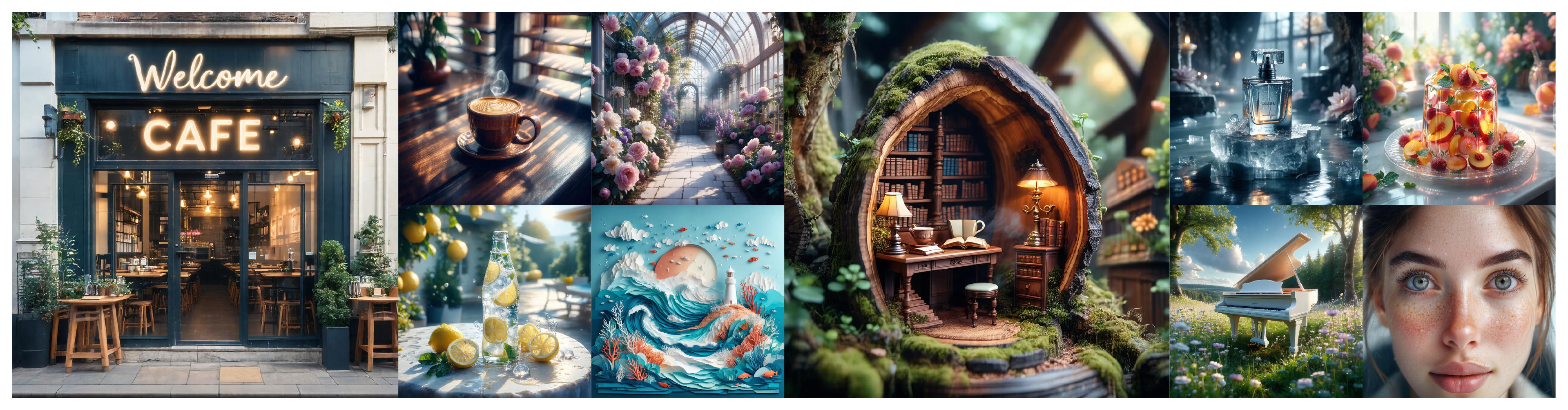} 
\caption{Sample images generated by \method{} on the SD 3.5-Large backbone. The generations align with the text prompt and human preference while keeping strong visual aesthetics.}
\label{fig:teaser}
\end{figure}

\section{Introduction}
\label{sec:intro}

Text-to-image flow and diffusion models fit a data distribution, not the objectives users care about~\cite{esser2024sd3,lipman2023flow}. For any given reward (compositional correctness, accurate text in the image, a preference score), the deployed model is generally untuned. Alignment post-training closes this gap, by reinforcement learning (RL) or preference optimization~\cite{black2024ddpo,wallace2024diffusiondpo,liu2025flowgrpo}, but the fix must be repeated for every checkpoint: rollouts, reward infrastructure, and training repeat for every model, reward, and version, at the scale of the model being aligned. Alignment therefore never accumulates; it restarts with each new checkpoint, and the most capable model in a family is, for any particular reward, typically the least aligned. 
Training-based alignment chooses the reward at training time and updates the model to follow it, and the same updated model then serves every test sample. Test-time alignment offers a different route: steer the frozen model during sampling toward the reward, without training and with the objective chosen at generation time, so the alignment can be task-specific and sample-specific~\cite{uehara2025inference}.

Test-time alignment methods differ in their supervision source, that is, in where the per-step signal comes from. Existing methods draw it from the reward function itself and inherit three structural costs. \emph{Timing:} a reward scores clean images, but a sampler needs supervision at every noise level, so the endpoint reward must be converted into a step-wise signal, by differentiating it through posterior estimates~\cite{chung2023dps}, evaluating it over many candidates~\cite{wu2023tds,singhal2025fk}, or training a value function for each model~\cite{li2024svdd}. \emph{Reach:} gradient-based conversion excludes non-differentiable rewards, and per-sample reward calls exclude rewards that are slow, human, or proprietary at deployment. \emph{Anchoring:} steering by the reward alone lacks the KL constraint that alignment training enforces, and over-optimizes~\cite{kim2025das}. The model stays frozen, but the supervision must be rebuilt at every sample, it excludes some rewards, and it carries no KL constraint.

An aligned model, kept together with its base, is a supervision source of a different kind. The model pair stores the training reward implicitly, as a density ratio~\cite{rafailov2023dpo}, and it stores more than the reward: alignment training shaped the whole trajectory distribution, so the pair also holds the reward's step-wise form at every noise level, pre-anchored by the training-time KL constraint and expressed in the sampler's own coordinates, as the method section makes precise. This is the object reward-based methods synthesize per step or train per model. The pair supplies it in two forward passes, without evaluating the reward function or its gradient. Language-model practice points the same way: decoding-time methods train dedicated reward models or score candidates with small pairs of tuned and untuned models~\cite{xu2025genarm,zhou2024weaktostrong}. In image generation, pair-form supervision exists publicly, but only the base model in the pair, which is rarely the one deployed, consumes it, to modulate the alignment strength~\cite{jin2025rlg}.

We explore whether model-form supervision can cross different models, especially model scales: can a small, weak model pair supervise the sampling of a larger, frozen model that was never tuned for the reward (test-time weak-to-strong alignment)? 
The extra computation at test time scales with the size of the models that supply the signal, so a pair much smaller than the host adds little. The aligned model that supplies this supervision need not be the model that generates, and need not be strong; it need only have absorbed the reward. If it can, training-time and test-time alignment become a pipeline: one RL run on a small \emph{source base model} from the same family returns its aligned counterpart, the
\emph{source aligned model}, and the two together form a \emph{source alignment pair}. Their implicit reward then steers the larger \emph{host base model} at test time, changing its sampling trajectory and leaving its weights untouched. Three observations make alignment supervision from a source alignment pair sufficient. The stored object is small: the KL constraint keeps the source aligned model near its base, so the pair holds only the change the alignment run made, a reward direction rather than a generative distribution. The pair contributes information, not capability: its difference cancels what two source models share, so a pair whose alignment run changed nothing leaves the host base model unchanged. The pair's estimate is used only where it is valid: on the support overlap, the pair evaluates states the host base model visits in-distribution. The transfer has a closed form: under the shared noising kernel, tilting the host base model by the source alignment pair's ratio displaces the sampling velocity by exactly the pair's velocity difference, a single additive term. Every time-dependent factor cancels, and the one-step tilted transition is realized exactly (Proposition~\ref{prop:transport}). The resulting method, \method{}, needs no noise-level schedule and no filtering; its single scalar controls the alignment strength and can extrapolate it beyond that of the source alignment pair. At test time the method adds two forward passes of the source alignment pair, and evaluates neither the reward nor its gradient, and alignment stays outside the host base model: its weights unmodified and controllable alignment strength per sample.

Empirically, the transfer holds across backbones and modalities. On Stable Diffusion~3.5, the guided host model improves over the host base model on every measured reward and can exceed the source aligned model itself (Table~\ref{tab:sd35_pickscore}); the same recipe carries over to a FLUX sibling and to text-to-video generation (Tables~\ref{tab:flux_krea} and~\ref{tab:wanfun}). Transfer behavior varies by reward in a predictable way: rewards whose signal forms at coarse spatial scale transfer at moderate strength while rewards that require fine low-noise structure need stronger guidance. 
 
Alignment training optimizes the objective once, so a checkpoint stays committed to the completed training process. \method{} gives the option of setting that objective at test time instead, on a host base model that stays untouched: which reward to follow, how strongly, and for which sample. Alignment therefore accumulates across a model family: one RL run yields supervision that upgrades siblings across scale and lineage at test time, and remains useful after any single checkpoint is replaced. Although a high-volume deployment with one fixed reward could distill the guided distribution back into the host base model, that route amortizes over volume, not over the product of many rewards, rotating checkpoints, and personalization. Our contributions:
\begin{itemize}
    \item We propose test-time weak-to-strong alignment, in which a source alignment pair (not a reward function) supervises the sampling of a larger, frozen host base model. The framework turns training-time and test-time alignment into a pipeline: one cheap RL run yields the source alignment pair, which then aligns any host base model sharing its latent space, larger or not, leaving its weights untouched.
    \item We derive the exact transport of the source alignment pair's implicit reward into the host base model's sampler: the density-ratio tilt reduces to one additive velocity term, all time dependence cancels, and the one-step tilted transition kernel is realized exactly (Proposition~\ref{prop:transport}). Two consistency limits fix every design choice, leaving no schedule and only a single scalar strength, and we state the remaining approximations explicitly. The velocity-difference form appears in prior work~\cite{jin2025rlg,gao2025delta,liu2024proxy}; the cross-scale anchor swap and its analysis are ours.
    \item Across image and video flow models (Stable Diffusion~3.5, FLUX, and Wan backbones), \method{} lifts the host base model on preference, compositional, and text-rendering rewards, can exceed the source aligned model itself, and generalizes across scale and across siblings of a family (Tables~\ref{tab:sd35_pickscore}--\ref{tab:wanfun}).  
\end{itemize}

\section{Related Work}
\label{sec:related}
\subsection{Diffusion Model Alignment via Post-Training}
Reward-based fine-tuning aligns diffusion and flow models with human objectives. DDPO~\cite{black2024ddpo} and DPOK~\cite{fan2023dpok} optimize rewards with policy gradients; Diffusion-DPO~\cite{wallace2024diffusiondpo} learns from preference pairs without an explicit reward model; DRaFT~\cite{clark2024draft} and AlignProp~\cite{prabhudesai2023alignprop} backpropagate differentiable rewards through sampling; Flow-GRPO~\cite{liu2025flowgrpo} and DanceGRPO~\cite{DanceGRPO} bring group-wise policy optimization to flow matching and produce the public checkpoint pairs we build on. Despite strong gains, post-training binds each reward to one checkpoint: the run must be repeated for every model, reward, and version, which motivates reuse rather than repetition.

\subsection{Test-Time Alignment of Diffusion Models}
Inference-time methods steer a frozen model toward a reward-tilted target~\cite{uehara2025inference}. Classifier and posterior guidance differentiate the reward through posterior estimates~\cite{dhariwal2021diffusion,chung2023dps}; twisted SMC and Feynman--Kac steering evaluate the reward over particle populations~\cite{wu2023tds,singhal2025fk}; soft value-based decoding trains a value function per model~\cite{li2024svdd}; direct noise optimization refines the initial noise~\cite{tang2024dno}; DAS regularizes against reward over-optimization~\cite{kim2025das}; DyMO schedules multi-objective guidance dynamically along the trajectory~\cite{xie2025dymo}; HyperAlign generates input-conditioned adapter weights with a hypernetwork~\cite{xie2026hyperalign}. All of these draw their supervision from the reward function, so each sample requires gradient, particle, or value-function access, and non-differentiable or deployment-restricted rewards remain out of reach. We instead draw test-time supervision from a trained model pair, in which the reward arrives pre-converted and pre-anchored.

\subsection{Weak-to-Strong Learning and Guidance}
Weak-to-strong generalization studies weak supervision of strong model training~\cite{burns2024weak}. At decoding, DExperts and contrastive decoding compose tuned--untuned pairs~\cite{liu2021dexperts,li2023contrastive}; proxy-tuning and emulated fine-tuning apply a small pair's logit offset to a larger model~\cite{liu2024proxy,mitchell2024emulator}; weak-to-strong search re-ranks candidates by a small pair's log-ratio~\cite{zhou2024weaktostrong}; GenARM trains a reward model for decoding-time guidance~\cite{xu2025genarm}; DeRa rescales alignment strength at decoding~\cite{liu2024dera}. In generation, autoguidance sharpens a strong model by pushing away from a degraded copy~\cite{karras2024autoguidance}; W2SD refines sample quality with weak--strong differences~\cite{bai2026w2sd}; RLG composes an RL pair with its own base to control that model's alignment strength~\cite{jin2025rlg}; Delta Sampling injects adapted-minus-base residuals into another model to transfer adapters and styles~\cite{gao2025delta}. Closer to our setting, MotionEcho corrects a distilled student's sampling trajectory with a stronger diffusion teacher at inference time, supervising a weak model with a strong one rather than the reverse~\cite{rong2026motionecho}, and implicit motion transfer moves motion from reference clips onto a static image~\cite{li2026fleximmt}. None of these transfers an RL-learned reward to a different, never-tuned model: discrete states force search or re-ranking, and existing guidance keeps the pair's signal inside the model that produced it or moves objects other than rewards. In flow models the same log-ratio tilt becomes a closed-form drift, which enables training-free, search-free reward transfer across scale.

\section{Method}
\label{sec:method}

\subsection{Preliminaries}
\label{sec:prelim}

\noindent\textbf{Flow-Matching Generative Models.}\enspace
Current generators~\cite{FLUX, wanfun1p3B,esser2024sd3} adopt the flow-matching formulation~\cite{lipman2023flow,liu2023rectified} to transport sampled Gaussian noise to clean data under the control of a condition $c$, typically a user prompt for image or video generation (omitted from the notation for brevity). With noise level $t$ running $1\to0$, training corrupts data through the forward noising kernel:
\begin{equation}
z_t=(1-t)\,z_0+t\,\epsilon,\qquad \epsilon\sim\mathcal N(0,I),
\label{eq:interp}
\end{equation}
and the network learns the velocity field by regression,
\begin{equation}
\min_\theta\;\mathbb E_{z_0,\epsilon,t}\big\|v_\theta(z_t,t)-(\epsilon-z_0)\big\|_2^2,
\label{eq:fmloss}
\end{equation}
so that $v_\theta(z,t)\approx\mathbb E[\epsilon-z_0\,|\,z_t=z]$. The velocity determines the marginal score $\nabla_z\log p_t(z)$ through the standard identity~\cite{song2021score,lipman2023flow}:
\begin{equation}
s_\theta(z,t)\;=\;-\,\frac{z+(1-t)\,v_\theta(z,t)}{t},
\label{eq:sv}
\end{equation}
whose coefficients depend only on $(z,t)$, not on the model. Generation runs a discretized sampler from $z_1\sim\mathcal N(0,I)$, each step moving the state from noise level $t$ to the next $t'$:
\begin{equation}
z_{t'}\;=\;a_t\,z_t\;+\;b_t\,v_\theta(z_t,t)\;+\;\sigma_t\,\varepsilon,
\label{eq:step}
\end{equation}
where the coefficients $a_t,b_t,\sigma_t$ come from the scheduler alone. We use the deterministic sampler (setting $\sigma_t{=}0$) by default.

\noindent\textbf{Alignment and Post-Training.}\enspace
A pretrained generator models the training data distribution. 
To improve how well it satisfies objectives such as adherence to the text prompt, compositional correctness, or human preference, alignment post-training maximizes a reward function $r(z_0)$ that measures the chosen objective on the clean output, while a KL term keeps the tuned model close to the pretrained distribution $p_\theta$~\cite{ziebart2008maximum,rafailov2023dpo}:
\begin{equation}
\max_{\theta_R}\;\;\mathbb E_{z_0\sim p_{\theta_R}}\big[r(z_0)\big]\;-\;\beta\,\mathrm{KL}\big(p_{\theta_R}(z_0)\,\big\|\,p_\theta(z_0)\big),
\label{eq:rlobj}
\end{equation}
whose optimum is the reward-tilted distribution $p_{\theta_R}(z_0)=\tfrac{1}{Z}\,p_\theta(z_0)\,e^{\,r(z_0)/\beta}$, with the normalization constant $Z$. The reward is evaluated on the decoded image; we omit the decoder from the notation. Rearranged, the optimum shows that the aligned model and its base implicitly store the training reward in a log-density ratio~\cite{rafailov2023dpo}:
\begin{equation}
\frac{r(z_0)}{\beta}=\log\frac{p_{\theta_R}(z_0)}{p_\theta(z_0)}+\mathrm{const}.
\label{eq:implicit}
\end{equation}

\subsection{Motivation and Problem Definition: From Reward to Model-Form Supervision}
\label{sec:problem}

Alignment post-training has become a dominant recipe for adapting current generators to target objectives. It updates the model parameters with supervision signals from a reward and the training data. Although this improves alignment overall, the whole parameter set is updated to satisfy all the training data at once, which may leave specific cases uncovered. Moreover, the cost of running Eq.~\eqref{eq:rlobj} is tied to both the model being trained and the reward, and a finished run remains bound to the parameters of its checkpoint. Once the base model is updated, the alignment does not carry over and must be redone. Thus, when a large flow-matching model $\theta_l$ serves as the deployed base model (\ie, the \emph{host base model}) and specific cases remain unaligned for a given objective and reward, re-running RL on it risks making post-training itself the bottleneck.

\noindent\textbf{Test-Time Alignment.}\enspace
Test-time alignment deploys the alignment to test time. It keeps the base model $\theta_l$ frozen and steers the sampling toward the reward-tilted distribution $\tfrac{1}{Z}\,p_l(z_0)\,e^{\,r(z_0)/\beta}$, the optimum of Eq.~\eqref{eq:rlobj} with $p_l$ in place of $p_\theta$~\cite{uehara2025inference}. 
The alignment objective and its effect can therefore be applied flexibly at inference time, for each specific case. 
The steering, however, needs a signal at every sampling step, and existing methods take this signal from the reward function itself. Gradient-based guidance differentiates $r$ on the one-step predicted clean sample $\hat z_0$~\cite{chung2023dps, xie2025dymo, kim2025das}, which requires a differentiable reward and a backward pass through $\theta_l$ at every step. 
Search-based methods avoid gradients by drawing many candidates and keeping those that $r$ scores highest~\cite{wu2023tds,singhal2025fk}, so a single sample costs many denoising runs and reward evaluations. Both read the reward on the predicted $\hat z_0$, and at high noise levels the prediction is blurry, so the early signal is unreliable. Value-based methods require training a step-wise surrogate for different models and rewards~\cite{li2024svdd}.

\noindent\textbf{Supervision from a Model Pair.}\enspace
The reward measures the generated image, so its signal and its gradient reach the sampler only through the trajectory that produced that image. 
Reward-based alignment also requires gradient calculation through the differentiable reward model.  
The test-time alignment methods spend test-time computation to obtain a per-step signal from a reward defined at the endpoint. In a post-training process, solving Eq.~\eqref{eq:rlobj} shapes the whole sampling trajectory, not only the clean output. Eq.~\eqref{eq:implicit} also shows that, relative to its base $\theta$, the aligned model $\theta_R$ holds the reward as a log-density ratio. Evaluating this implicit reward takes only forward passes of the two models, with no reward function, no gradient, and no surrogate training. The ratio is defined at every noise level, so the supervision is dense rather than available only on the clean output. 
The pair need not be the deployed host model: the reward signal can come from another model, especially a small model trained by RL at low cost~\cite{liu2025flowgrpo,zheng2026diffusionnft}, even though the host model $\theta_l$ to be aligned at test time is large. 
Such an implicit reward has been used only within the pair, to modulate the alignment strength of the pair's own base model~\cite{jin2025rlg}. In this paper, we transfer the implicit reward of a small pair to the sampling of the large frozen model $\theta_l$, so that one cheap RL run on a weak model aligns the strong model at test time.

\subsection{\method{}: Weak-to-Strong Guidance} 
\label{sec:ours}

\noindent\textbf{Implicit-reward transfer.}\enspace
As discussed in Section~\ref{sec:problem}, we construct \emph{test-time weak-to-strong alignment} by replacing the test-time supervision with a trained model pair, combining the strengths of both routes: run the RL once on a small (weak) model, and transfer the learned alignment to the large (strong) host base model at test time, with its weights frozen. Specifically, a \emph{source base model} $\theta_s$ (\ie, the small model before RL) from the same family is post-trained against $r$ by Eq.~\eqref{eq:rlobj} into the \emph{source aligned model} $\theta_s^{\mathrm{RL}}$ (\ie, the same model after RL), and such pairs are often publicly released~\cite{liu2025flowgrpo,DanceGRPO}. Kept two source models together as the \emph{source alignment pair}, the \emph{implicit reward} of Eq.~\eqref{eq:implicit} can be obtained: the reward becomes available in model form, and evaluating it requires only forward passes. Throughout, the three models share the latent space, the conditioning encoders, and the forward noising kernel Eq.~\eqref{eq:interp}, as members of one model family do.

Substituting the implicit reward into the reward-tilted distribution of the host base model gives the target of \method{}:
\begin{equation}
p_l^{\star}(z_0)\;\propto\;p_l(z_0)\,e^{\,w\,r(z_0)/\beta}\;\propto\;p_l(z_0)\left[\frac{p_s^{\mathrm{RL}}(z_0)}{p_s(z_0)}\right]^{w},
\label{eq:target}
\end{equation}
where $r(z_0)/\beta$ in the exponent is exactly the implicit reward of Eq.~\eqref{eq:implicit}: the source alignment pair supplies it as the log-ratio $\log\big(p_s^{\mathrm{RL}}(z_0)/p_s(z_0)\big)$, which stores the preferences of the training reward. The target therefore involves only the three frozen models and never queries the reward function $r$. The scalar $w>0$ controls the KL penalty, where larger $w$ tilts the distribution further toward the reward. Taking log-density gradients of the target splits it into the host base model's score plus the pair's score difference,
\begin{equation}
\nabla_{z_0}\log p_l^{\star}(z_0)\;=\;\nabla_{z_0}\log p_l(z_0)\;+\;w\,\big(\nabla_{z_0}\log p_s^{\mathrm{RL}}(z_0)-\nabla_{z_0}\log p_s(z_0)\big).
\label{eq:score}
\end{equation}

\begin{figure}[t]
\centering
\begin{minipage}[t]{0.52\textwidth}
  \vspace{0pt}
  \centering
  \includegraphics[width=\linewidth]{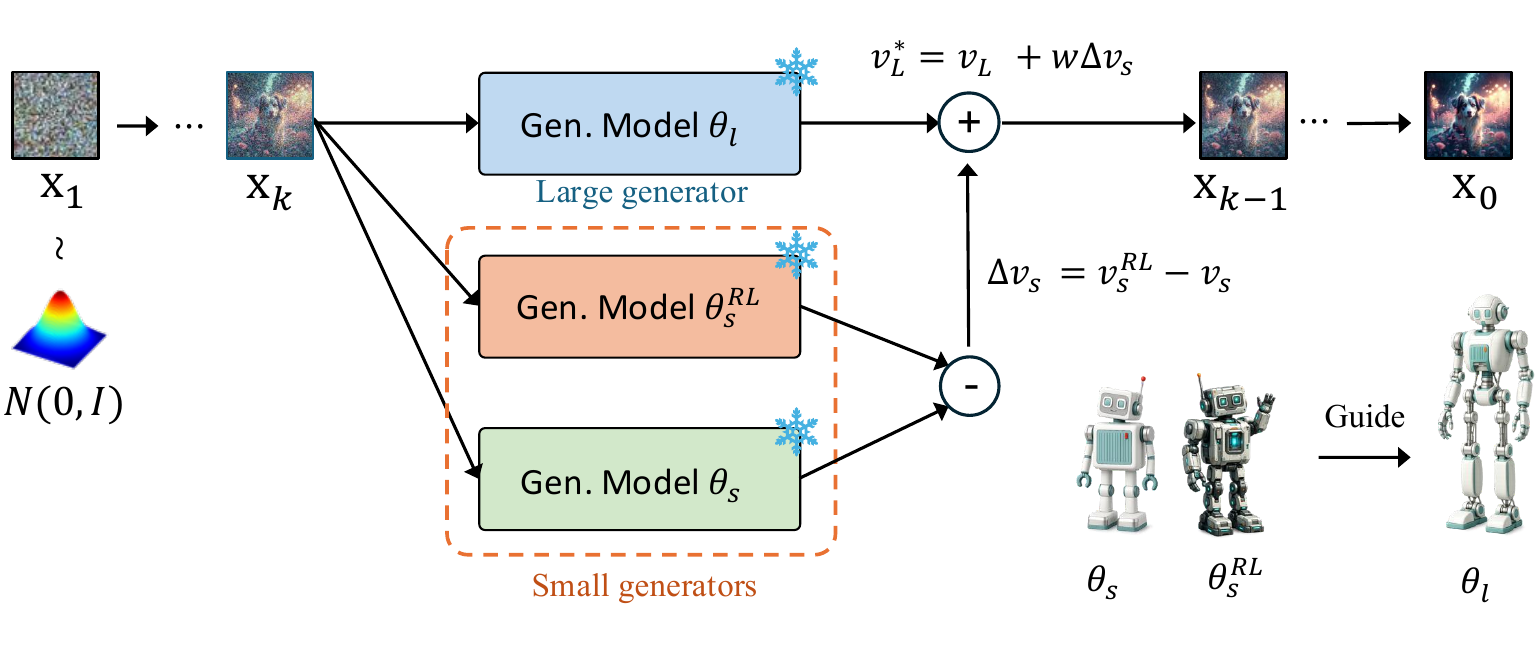}
  \caption{Test-time weak-to-strong alignment. At every sampling step, the source alignment pair $(\theta_s,\theta_s^{\mathrm{RL}})$ supplies the alignment delta $\Delta v_s$, and the host base model $\theta_l$ adds it to its own velocity; the sampler is otherwise unchanged.}
  \label{fig:framework}
\end{minipage}
\hfill
\begin{minipage}[t]{0.44\textwidth}
  \setlength{\parskip}{0pt}%
  \vspace{5mm}
  \hrule height.8pt\kern3pt
  \captionsetup{singlelinecheck=false,skip=0pt,belowskip=0pt}
  \captionof{algorithm}{\method{}}%
  \label{alg:aligngraft}\par
  \kern3pt\hrule\kern4pt
  \small
  \begin{algorithmic}[1]
  \REQUIRE host base model $\theta_l$, source alignment pair $(\theta_s,\theta_s^{\mathrm{RL}})$, prompt $c$, strength $w$, schedule $\{t_k\}_{k=1}^{N+1}$
  \STATE $z\sim\mathcal N(0,I)$
  \FOR{$k=1$ \TO $N$}
      \STATE read $v_l,\,v_s^{\mathrm{RL}},\,v_s$ at $(z,t_k,c)$, each with its own CFG scale
      \STATE $v_l^{\star}\gets v_l+w\,(v_s^{\mathrm{RL}}-v_s)$ \hfill$\triangleright$ the only method line
      \STATE $z\gets a_{t_k}z+b_{t_k}v_l^{\star}+\sigma_{t_k}\varepsilon$,\quad $\varepsilon\sim\mathcal N(0,I)$
  \ENDFOR
  \RETURN $\mathrm{decode}(z)$
  \end{algorithmic}
  \kern2pt\hrule height.8pt
\end{minipage}
\end{figure}

The resulting method is simple: at each sampling step, we read all three frozen models at the same state, form the \emph{alignment delta} from the source alignment pair, add it to the host base model's velocity, and take the ordinary sampler step (Figure~\ref{fig:framework}, Algorithm~\ref{alg:aligngraft}). Nothing is trained, no weights are merged, and the reward never appears.

\noindent\textbf{Alignment delta from the source alignment pair.}\enspace
The decomposition in Eq.~\eqref{eq:score} is stated on clean data and in scores, while the sampler runs on noisy states and consumes velocities. Because the models share the forward noising kernel, the same decomposition holds at every noise level, with the noisy marginal $p_{l,t}(z)$ tilted by the ratio $\big[p^{\mathrm{RL}}_{s,t}(z)/p_{s,t}(z)\big]^{w}$; substituting the identity Eq.~\eqref{eq:sv} into this decomposition then cancels every $t$-dependent factor (Proposition~\ref{prop:transport}), and the pair's term simplifies to a velocity difference, the \emph{alignment delta}:
\begin{equation}
\Delta v_s(z,t)\;=\;v_s^{\mathrm{RL}}(z,t)-v_s(z,t).
\label{eq:delta}
\end{equation}
The subtraction cancels everything the two source models share (their content, biases, and limited capability) and keeps only what alignment changed: the reward direction. A pair that learned nothing changes nothing. Moreover, the delta is the right object at \emph{every} noise level: because alignment training shaped the whole trajectory distribution, the ratio of the pair's noisy marginals equals the value function of the source alignment pair's process at each $t$ (derivation in the appendix), so the pair provides exactly the step-wise supervision that reward-based methods must synthesize or train. Consequently the reward needs no gradient and no test-time access, and the interface is agnostic to how the source aligned model was obtained: online RL or preference optimization define the same ratio~\cite{liu2025flowgrpo,wallace2024diffusiondpo}. No adapter weights are transferred or applied to the host base model.

\noindent\textbf{Guided velocity.}\enspace
Adding the weighted alignment delta to the host base model's velocity gives the sampling rule of \method{}:
\begin{equation}
v_l^{\star}(z,t)\;=\;v_l(z,t)\;+\;w\,\Delta v_s(z,t),
\label{eq:rule}
\end{equation}
where $w$ carries over from the target Eq.~\eqref{eq:target}; it is the method's only hyperparameter, swept in Section~\ref{sec:strength}. Because the cancellation is exact, no noise-level schedule, window, or filter is needed, and the same $w$ applies along the whole trajectory. The target also factorizes symmetrically, $p_l\cdot(p_s^{\mathrm{RL}}/p_s)=p_s^{\mathrm{RL}}\cdot(p_l/p_s)$: the same distribution is the host base model tilted by the source alignment pair's reward, or the source aligned model tilted by the capability gap between large and small. The second reading makes the method weak-to-strong and locates its boundary: the ratio is well-defined only where the models' supports overlap.

\noindent\textbf{Sampling.}\enspace
Each model is read through its own deployed sampler, with classifier-free guidance composed at its trained scale. The guided velocity $v_l^{\star}$ then enters the unmodified sampler step Eq.~\eqref{eq:step}. Two consistency limits double as bit-exact self-checks: if the host base model equals the source base model, \method{} reproduces the source aligned model exactly; if the pair is null ($\theta_s^{\mathrm{RL}}{=}\theta_s$), the host base model is returned unchanged. Cost: three forward passes per step instead of one (a small constant overhead since the pair is small), with zero backward passes and zero reward evaluations; in our instantiation the source aligned model is a LoRA over $\theta_s$, so the pair also shares memory.

\begin{figure}[t]
  \centering
  \includegraphics[width=1.0\linewidth]{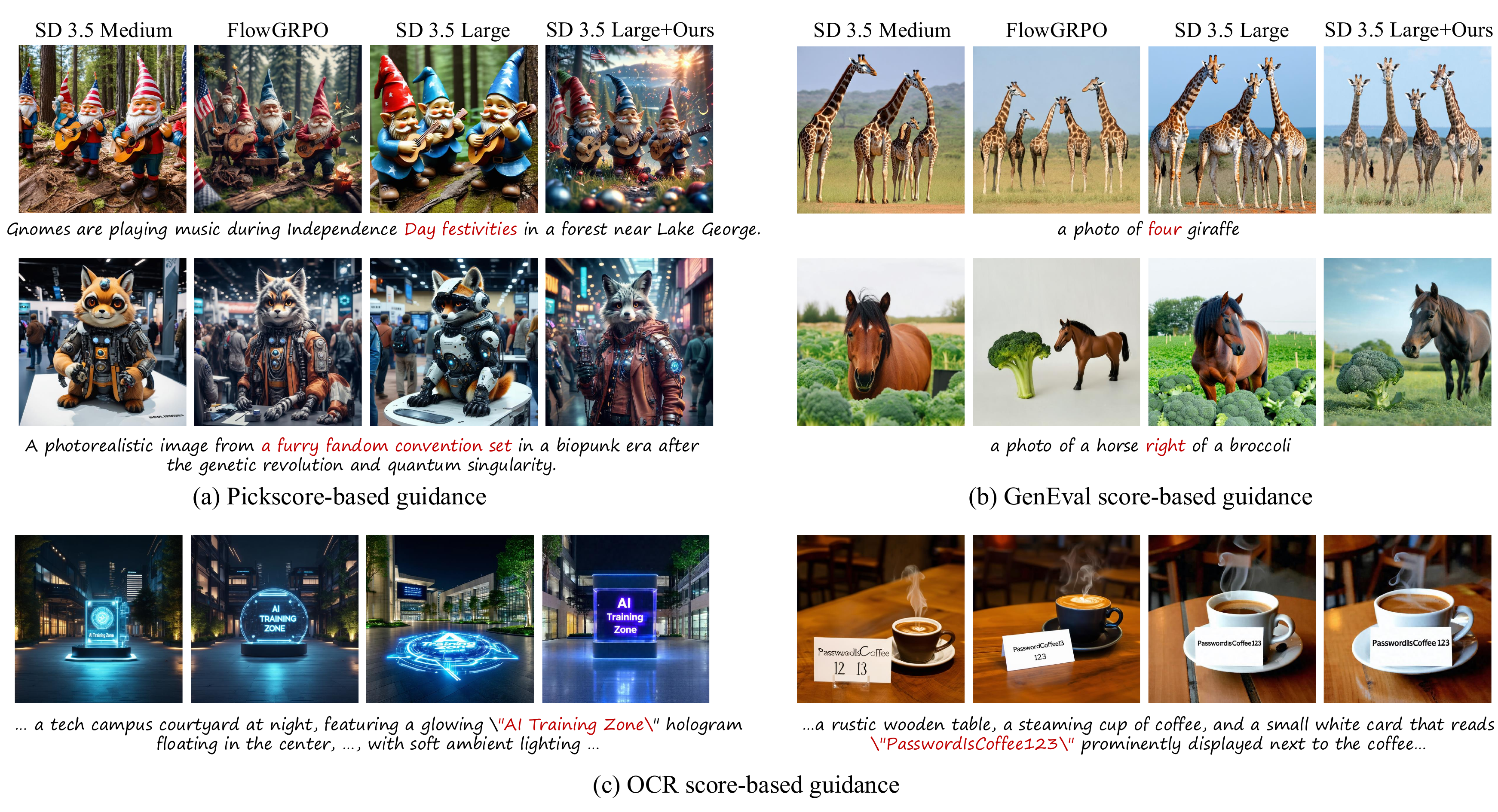}
      \caption{Qualitative transfer of three reward sources on SD 3.5, in blocks (a) PickScore, (b) GenEval, and (c) OCR. Columns, left to right: source base (SD 3.5-Medium), source aligned (Flow-GRPO), host base (SD 3.5-Large), and guided host (SD 3.5-Large\,+\,\method{}). Each reward transfers its own
      behavior while the host base model keeps its native fidelity.}
\label{fig:teachers}
\end{figure}

\noindent\textbf{Why the transfer is exact.}\enspace
Eq.~\eqref{eq:rule} is derived from the tilted score at a single noise level. We then ask whether the substitution is exact or only a first-order approximation, and whether it survives the sampler's discretization. The proposition below answers both, for the marginal and for one sampler step.
\begin{proposition}[exact transport]
\label{prop:transport}
Under the shared forward kernel: (i) tilting the host base model's marginal at noise level $t$ by the pair's density ratio raised to $w$ shifts the velocity by exactly $w\,\Delta v_s(z,t)$, with all $t$-dependent factors cancelling; and (ii) for the stochastic sampler step Eq.~\eqref{eq:step} with $\sigma_t{>}0$, taken from a shared state, sampling with $v_l^{\star}$ realizes exactly the reward-tilted one-step transition of the host base model. (Proof in the appendix.)
\end{proposition}  
\noindent
The proposition rules out two alternatives: any scheme that reweights $\Delta v_s$ by a function of $t$ implicitly denies that the tilt is a density ratio, and any linearization of the step is unnecessary. Part~(i) holds for every $\sigma_t$; part~(ii) needs $\sigma_t{>}0$. Our deterministic sampler is the limit $\sigma_t{\to}0$, where part~(ii) reduces to part~(i). Two approximations remain, and we state them: the per-noise-level tilted family is the standard product-of-experts construction (the same status classifier-free guidance has~\cite{bradley2024cfg}) and the source alignment pair's value function stands in for the host base model's, which is valid on the support overlap of the models; both are discussed in the appendix.

\begin{figure}[t]
  \centering
  \includegraphics[width=1.0\linewidth]{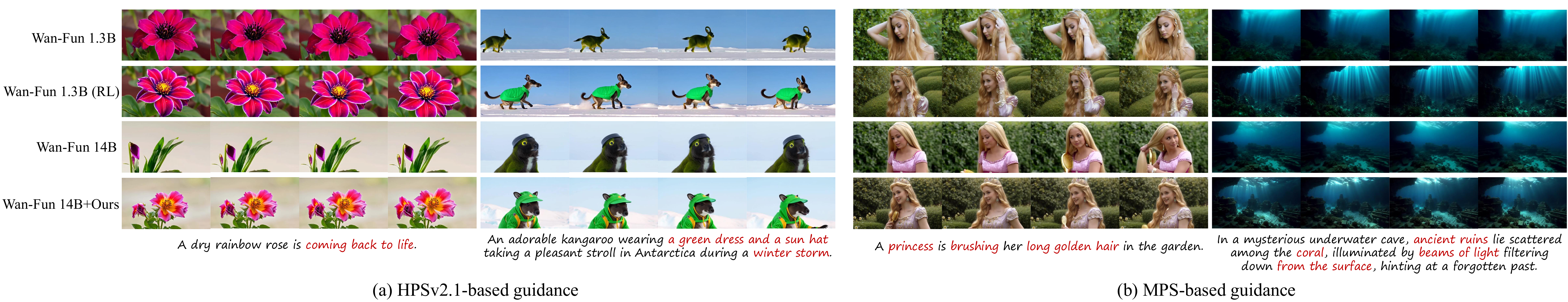}
      \caption{Qualitative comparison of text-to-video transfer on Wan-Fun, with (a) HPS v2.1 and (b) MPS reward sources. Rows, top to bottom: source base (Wan-Fun-1.3B), source aligned (Wan-Fun-1.3B RL), host base (Wan-Fun-14B), and guided host (Wan-Fun-14B\,+\,\method{}); each row shows four frames of a clip. The transfer improves per-frame appeal and prompt adherence while preserving motion (Table~\ref{tab:wanfun}).}
\label{fig:wan}
\end{figure}

\begin{figure}[t]
  \centering
  \includegraphics[width=1.0\linewidth]{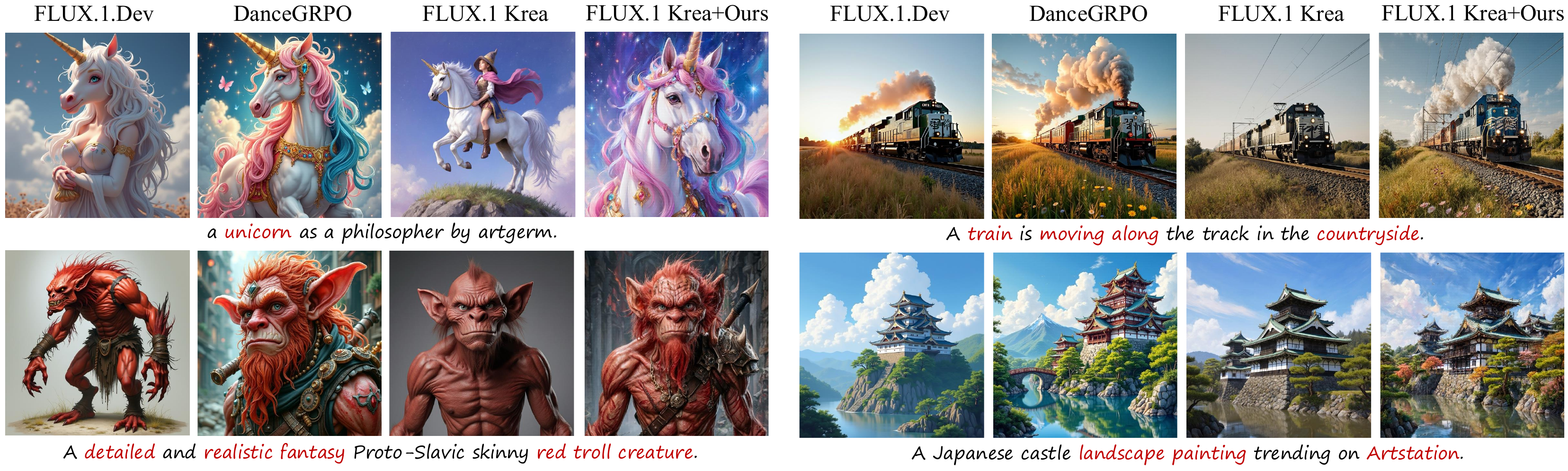}
  \caption{Qualitative comparison of cross-sibling transfer within the FLUX family.}
\label{fig:flux}
\end{figure}

\section{Experiments}
\label{sec:experiments}

We evaluate \method{} across generative backbones and domains, validating it through multi-dimensional comparisons: quantitative and qualitative results, generalization across model families and modalities, and analyses of guidance strength and design choices.

\subsection{Experimental Setting}
\label{sec:setup}
\noindent\textbf{Implementation details.}\enspace
For text-to-image (T2I) generation, we use the 2.5B SD 3.5-Medium~\cite{esser2024sd3} as the source base, paired with its post-RL counterparts trained by Flow-GRPO~\cite{liu2025flowgrpo} with PickScore~\cite{kirstain2023pickapic}, OCR~\cite{ocr}, and GenEval~\cite{ghosh2023geneval} rewards, to guide the frozen 8B SD 3.5-Large~\cite{esser2024sd3}; the family shares one latent space and text encoders~\cite{stabilityai2024sd35}. We further use FLUX.1-dev~\cite{FLUX} as the source base, paired with its DanceGRPO counterpart trained on the HPS v2.1 reward~\cite{DanceGRPO,wu2023hpsv2}, to enhance the frozen FLUX.1-Krea~\cite{FLUX_kera}. For text-to-video (T2V) generation, we use Wan-Fun-1.3B~\cite{wanfun1p3B} as the source base, paired with reward-aligned counterparts trained by reward backpropagation with the HPS v2.1 and MPS rewards~\cite{wu2023hpsv2,zhang2024mps,wanfunreward}, to enhance the frozen Wan-Fun-14B~\cite{wanfun14B}. All experiments use one NVIDIA H100 GPU for the SD 3.5 backbones and one B300 GPU for the FLUX and Wan backbones.

\noindent\textbf{Datasets and metrics.}\enspace
For T2I generation we evaluate on all 553 prompts from GenEval (four seeds per prompt)~\cite{ghosh2023geneval}, 400 prompts from HPD~\cite{wu2023hpsv2}, and 1K prompts from the OCR test set~\cite{liu2025flowgrpo}. For T2V generation we evaluate on 100 prompts sampled from MovieGenBench~\cite{polyak2024movie}. Automatic evaluators assess quality: PickScore~\cite{kirstain2023pickapic}, ImageReward (IR)~\cite{xu2023imagereward}, HPS v2.1~\cite{wu2023hpsv2}, and MPS~\cite{zhang2024mps} for human preference, the OCR scorer~\cite{ocr} for text rendering, the GenEval scorer~\cite{ghosh2023geneval} for prompt alignment, and the Aesthetic predictor~\cite{Aesthetic} for visual appeal; for video we additionally report HPSv3~\cite{ma2025hpsv3} and the visual-quality (VQ), motion-quality (MQ), and text-alignment (TA) scores of VideoAlign~\cite{liu2025videoalign}. Higher is better for all metrics.

\begin{table}[t]
\centering
\begin{minipage}[t]{0.49\textwidth}
\centering
\small
\setlength{\tabcolsep}{2pt}
\begin{tabular}{@{}lcccc@{}}
\toprule
Method & Aes & Pick & IR & HPS \\
\midrule
SD 3.5-M (source base) & 5.929 & 22.54 & 1.084 & 0.3003 \\
Flow-GRPO (source aligned)    & 6.350 & 23.85 & \textbf{1.404} & 0.3318 \\
\midrule
SD 3.5-L (host base)       & 5.982 & 22.70 & 1.154 & 0.3072 \\
\rowcolor{tabcolor} \hlfirst{SD 3.5-L + \method{}}  & \textbf{6.496} & \textbf{24.06} & 1.384 & \hllast{\textbf{0.3360}} \\
\bottomrule
\end{tabular}
\caption{PickScore reward source on SD 3.5 ($w{=}3$). The guided host improves on every metric and exceeds the source aligned model.}
\label{tab:sd35_pickscore}
\end{minipage}
\hfill
\begin{minipage}[t]{0.49\textwidth}
\centering
\small
\setlength{\tabcolsep}{3pt}
\begin{tabular}{@{}lcccc@{}}
\toprule
Method & Aes & Pick & IR & HPS \\
\midrule
FLUX.1-dev (source base) & 6.217 & 22.75 & 1.097 & 0.3087 \\
DanceGRPO (source aligned)       & 6.445 & 23.12 & 1.276 & 0.3463 \\
\midrule
FLUX.1-Krea (host base)      & 6.041 & 22.86 & 1.143 & 0.3051 \\
\rowcolor{tabcolor} \hlfirst{FLUX.1-Krea + \method{}}  & 6.220 & 23.10 & 1.229 & \hllast{0.3314} \\
\bottomrule
\end{tabular}
\caption{HPS v2.1 reward source transferred between equally sized FLUX siblings ($w{=}1$), which isolates lineage from scale. The guided host improves on all four metrics.}
\label{tab:flux_krea}
\end{minipage}
\end{table}

\subsection{Qualitative Evaluation}
\label{sec:qual}
Figure~\ref{fig:teachers} shows that the transfer is effective under different rewards we consider. For all three source alignment pairs (PickScore, GenEval, and OCR), guiding the frozen SD 3.5-Large host moves its samples toward the reward that source aligned model was trained on, and each source transfers a recognizably different behavior: the PickScore pair yields richer lighting, texture, and subject appeal; the GenEval pair corrects compositional structure such as object count and left/right arrangement that the host otherwise misses; and the OCR pair produces legible in-image text where the unguided host renders malformed glyphs. In every case the transferred sample keeps the host base model's native fidelity and resolution, so the pair contributes a reward-aligned direction rather than overwriting the host's prior. Consistent with the distributional nature of the transfer, the effect is a shift in success rate across prompts rather than a guarantee on individual seeds.

\subsection{Quantitative Analysis}
\label{sec:quant}

\begin{table}[t]
\centering
\small 
\setlength{\tabcolsep}{8.5pt}
\begin{tabular}{@{}lccccccc@{}}
\toprule
Method & Overall & Single obj. & Two obj. & Counting & Colors & Position & Color attr. \\
\midrule
SD 3.5-M (source base) & 0.63 & 0.98 & 0.82 & 0.55 & 0.81 & 0.28 & 0.52 \\
Flow-GRPO (source aligned)    & 0.93 & 0.99 & 0.98 & 0.88 & 0.91 & 0.90 & 0.82 \\
\midrule
SD 3.5-L (host base)       & 0.68 & 0.97 & 0.85 & 0.67 & 0.84 & 0.28 & 0.61 \\
\rowcolor{tabcolor} \hlfirst{SD 3.5-L + \method{}}  & 0.90 & 0.99 & 0.97 & 0.84 & 0.87 & 0.91 & \hllast{0.79} \\
\bottomrule 
\end{tabular}
\caption{GenEval reward source on SD 3.5 ($w{=}2$). Every sub-task improves, most strongly position and counting.}
\label{tab:sd35_geneval}
\end{table}

\begin{table}[h]
\centering
\small
\setlength{\tabcolsep}{3.9pt}
\begin{tabular}{@{}llcccccccc@{}}
\toprule
Reward & Method & Aes & MPS & IR & HPS v2.1 & HPSv3 & VQ & MQ & TA \\
\midrule
\multirow{4}{*}{HPS v2.1}
& Wan-Fun-1.3B (source base) & 4.942 & 0.088 & $-$0.466 & 0.2235 & 2.887 & $-$0.829 & $-$0.500 & $-$0.962 \\
& Fun-Reward-HPS (source aligned) & 5.508 & 0.104 & 0.394 & 0.2778 & 7.865 & $-$0.403 & $-$0.430 & $-$0.430 \\
\cmidrule{2-10}
& Wan-Fun-14B (host base) & 5.029 & 0.095 & $-$0.168 & 0.2292 & 4.470 & $-$0.761 & $-$0.339 & $-$0.602 \\
\rowcolor{tabcolor} \cellcolor{white} & Wan-Fun-14B + \method{} & \textbf{5.533} & \textbf{0.111} & \textbf{0.549} & \textbf{0.2839} & \textbf{8.780} & \textbf{$-$0.314} & \textbf{$-$0.318} & \hllast{\textbf{$-$0.147}} \\
\midrule
\multirow{4}{*}{MPS}
& Wan-Fun-1.3B (source base) & 4.932 & 0.087 & $-$0.495 & 0.2171 & 1.743 & $-$0.818 & $-$0.464 & $-$0.894 \\
& Fun-Reward-MPS (source aligned) & 5.384 & 0.101 & $-$0.020 & 0.2510 & 5.615 & $-$0.468 & $-$0.399 & $-$0.580 \\
\cmidrule{2-10}
& Wan-Fun-14B (host base) & 5.187 & 0.097 & 0.060 & 0.2376 & 5.416 & $-$0.717 & $-$0.348 & $-$0.483 \\
\rowcolor{tabcolor} \cellcolor{white} & Wan-Fun-14B + \method{} & \textbf{5.530} & \textbf{0.112} & \textbf{0.500} & \textbf{0.2666} & \textbf{7.983} & \textbf{$-$0.331} & \textbf{$-$0.321} & \hllast{\textbf{$-$0.231}} \\
\bottomrule
\end{tabular}
\caption{Text-to-video transfer on Wan-Fun with the HPS v2.1 and MPS reward sources ($w{=}1$). The guided host exceeds both the host base model and the source aligned model on different metrics, with motion quality preserved. The HPS v2.1 and MPS rows are sampled with different random seeds.}
\label{tab:wanfun}
\end{table}

We mainly compare three reference points at matched sampling: the host base model, the source aligned model, and the guided host base model. Table~\ref{tab:sd35_pickscore} reports the PickScore source on SD 3.5: the transfer improves every metric over the host base model and exceeds the source aligned model on Aesthetic, PickScore, and HPS v2.1, so the transferred alignment, rendered at the host's capability, exceeds what the source aligned model reaches on its own. Table~\ref{tab:sd35_geneval} repeats the protocol for the compositional GenEval source: the transfer raises the host base's overall GenEval score from $0.68$ to $0.90$, recovering most of the gap to the source aligned model ($0.93$), with the largest gains on position ($0.28\!\rightarrow\!0.91$) and counting ($0.67\!\rightarrow\!0.84$).

\begin{figure}[!t]
  \centering
  \includegraphics[width=1.0\linewidth]{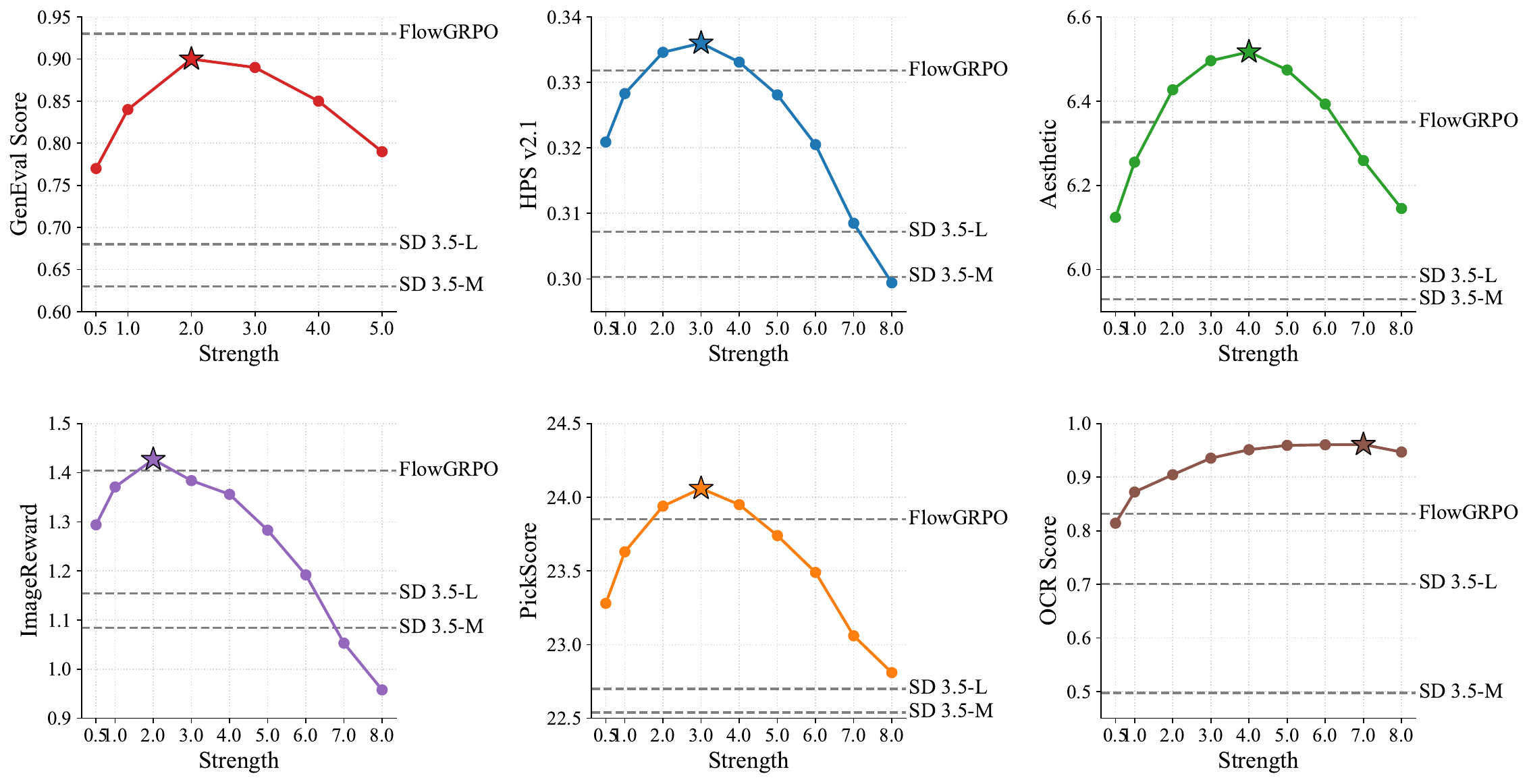}
      \caption{Guidance-strength search on SD 3.5: six reward metrics against $w\in[0.5,8]$, and $w\in[0.5,5]$ for GenEval, with source base, source aligned, and host base as dashed references. Each metric is scored under its own source alignment pair (GenEval and OCR under theirs; Aes, Pick, IR, HPS v2.1 under PickScore). Five metrics follow an inverted U, while the OCR reward rises to a plateau; every metric peaks above the source aligned model except GenEval, which peaks below it.}
\label{fig:search}
\end{figure}

\subsection{Generalization across Backbones and Modalities}
\label{sec:general}

The method requires only that the models share the latent space and the forward kernel; nothing restricts the host to a scaled-up member of the source lineage, or the data to images. We therefore examine two further settings: a sibling host within the same family, and the video modality.

\noindent\textbf{Same-family transfer (FLUX).}\enspace
In the SD 3.5 experiments the host base model is the scaled-up version of the source base, so capability and lineage change together. The FLUX setting isolates the lineage axis: FLUX.1-Krea is an equally sized sibling of the source base FLUX.1-dev that satisfies the shared-space assumption, yet is an independently post-trained checkpoint and was never tuned for the reward. As shown in Table~\ref{tab:flux_krea}, adding the DanceGRPO pair's alignment delta at $w{=}1$ improves the frozen Krea model on all four metrics and recovers $70\%$ of the source HPS gain, with consistent gains across prompt styles (Figure~\ref{fig:flux}). The implicit reward is thus attached to the shared latent space rather than to the checkpoint that produced it: any family member whose states the source alignment pair can evaluate in-distribution can consume the same delta.

\noindent\textbf{Video-domain transfer (Wan).}\enspace
The sampling rule Eq.~\eqref{eq:rule} carries over to text-to-video generation without modification, the latents now being spatio-temporal. Table~\ref{tab:wanfun} reports Wan-Fun backbones with two reward sources. With the HPS v2.1 source, guiding the frozen 14B host at $w{=}1$ improves HPS from $0.2292$ to $0.2839$, almost recovering the source gain; the MPS source behaves similarly (MPS $+0.015$, $107\%$). In both settings all held-out judges improve together, and the motion-quality score stays unchanged, so the gain is not obtained by trading motion dynamics for frame-wise appeal (Figure~\ref{fig:wan}). The Wan source alignment pairs come from reward backpropagation rather than policy-gradient RL. The interface accommodates this, because the sampling rule reads the pair's density ratio and never invokes $r$. Eq.~\eqref{eq:implicit} is the one step that needs the KL constraint: reward backpropagation anchors the pair by early stopping instead, so here we read $\Delta v_s$ as a reward direction rather than a calibrated $r/\beta$. The transfer holds regardless (Table~\ref{tab:wanfun}), which suggests the mechanism tolerates more than the derivation requires.

\subsection{Guidance Strength}
\label{sec:strength}
\begin{wrapfigure}{r}{0.5\linewidth}
\vspace{-1.2cm}
\centering
\includegraphics[width=1.0\linewidth]{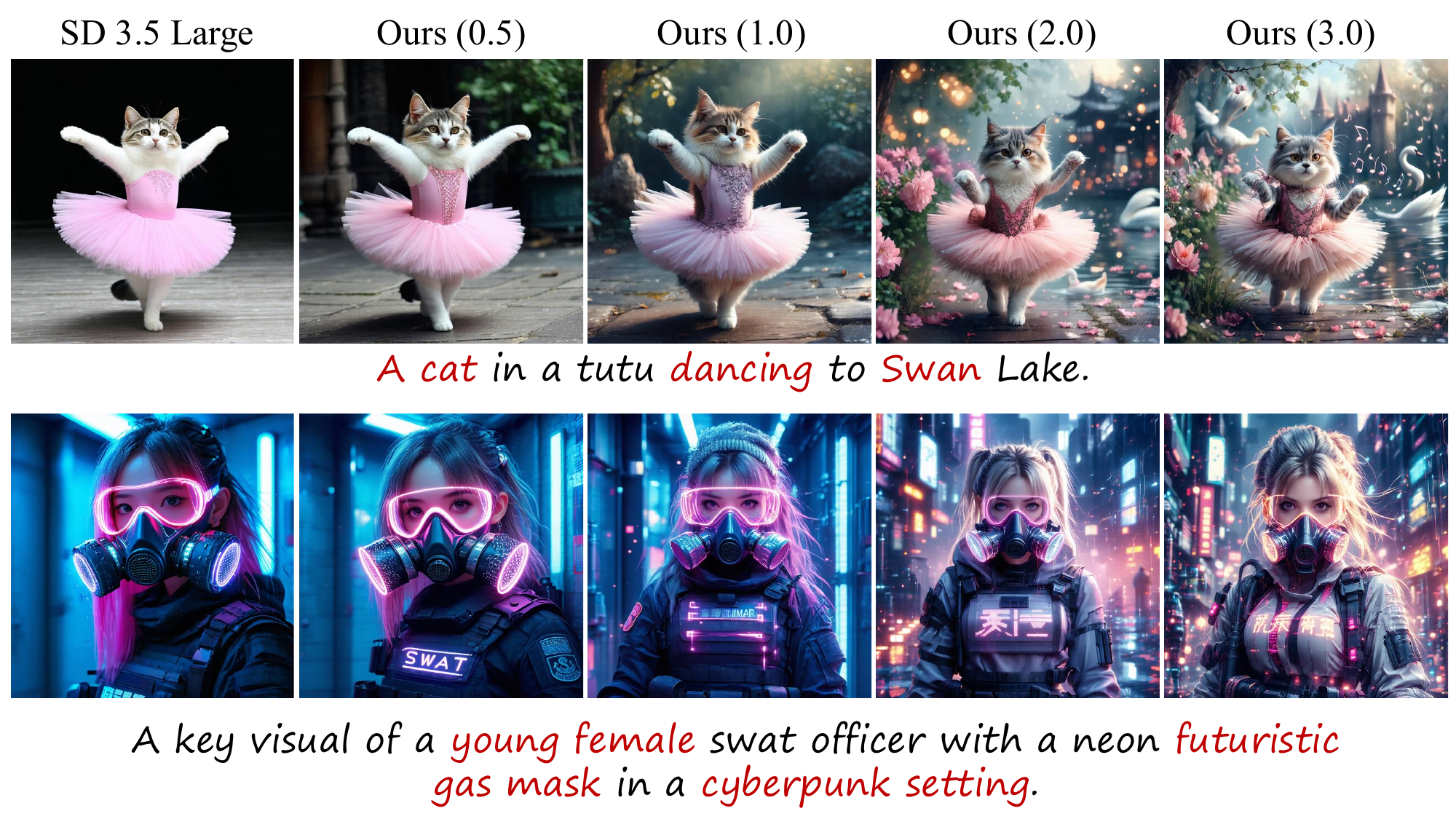}
\caption{\method{} under different guidance strength $w$ on SD 3.5 backbones.}
\label{fig:strength}
\vspace{-0.8cm}
\end{wrapfigure}
Figure~\ref{fig:search} sweeps $w$ quantitatively and Figure~\ref{fig:strength} shows the qualitative effect. We make three observations. First, every metric follows an inverted U: the reward rises to a broad peak at moderate strength and degrades beyond it. This decline is the measured onset of over-optimization, which bounds how far the strength can usefully be extrapolated. Second, at its peak the transfer matches or exceeds the source aligned model on most metrics while retaining the host's capability. Third, the peak location varies systematically by reward: preference and compositional rewards peak around $w\in[2,4]$.

\section{Conclusion}
\label{sec:conclusion}
We asked whether a small aligned model, kept with its base, can replace the reward function as the supervision source for test-time alignment, and answered with \method{}: a training-free transport of the source alignment pair's implicit reward into a frozen host base model's sampler, exact under the shared noising kernel. The reward never appears at generation time and need not be differentiable; the cost is two extra forward passes of the source alignment pair. The transfer is consistently effective: on every backbone we test, spanning scale, siblings of a family, and image and video generation, it improves the host base model on all reported metrics, recovers most of the source alignment gain, and at times exceeds the source aligned model itself. One RL run yields supervision that a whole model family can reuse; extending the transport beyond a shared latent space is left to future work.

\clearpage
\bibliographystyle{splncs04}
\bibliography{main}

\clearpage
\newpage
\beginappendix

\section{Score--Velocity Identity}
With the interpolation $z_t=(1-t)z_0+t\epsilon$ of the main text, the conditional distribution of $z_t$ given $z_0$ is $\mathcal N\big((1-t)z_0,\,t^2I\big)$. Tweedie's formula gives $\mathbb E[\epsilon\,|\,z_t=z]=-t\,s(z,t)$ and $\mathbb E[z_0\,|\,z_t=z]=\big(z+t^2 s(z,t)\big)/(1-t)$, where $s(z,t)=\nabla_z\log p_t(z)$ is the marginal score~\cite{song2021score}. Subtracting,
\begin{equation}
\begin{aligned}
v(z,t)&=\mathbb E[\epsilon-z_0\,|\,z_t=z]=-\frac{z+t\,s(z,t)}{1-t},\\
s(z,t)&=-\frac{z+(1-t)\,v(z,t)}{t}.
\end{aligned}
\label{eq:sv_app}
\end{equation}
Velocity and score are affine transforms of one another, with coefficients that depend only on $(z,t)$, not on the model. This is the fact that makes outputs of different models addable, and it is shared by all three models under the shared forward kernel.

\section{Sampler Coefficients}
The marginal-preserving SDE Euler step of Flow-GRPO~\cite{liu2025flowgrpo}, used unmodified in all experiments, is
\begin{equation}
z_{t'}=z_t\Big(1+\tfrac{\hat\eta^2}{2t}\,dt\Big)+v\,\Big(1+\tfrac{\hat\eta^2(1-t)}{2t}\Big)dt+\hat\eta\sqrt{-dt}\,\varepsilon,
\label{eq:sde_app}
\end{equation}
with $dt=t'-t<0$, $\hat\eta=\eta\sqrt{t/(1-t)}$, and $\varepsilon\sim\mathcal N(0,I)$. In the generic form of the main text, $a_t=1+\hat\eta^2 dt/(2t)$, $b_t=\big(1+\hat\eta^2(1-t)/(2t)\big)dt$, and $\sigma_t=\hat\eta\sqrt{-dt}$; all three depend only on the scheduler. Setting $\eta=0$ recovers the deterministic ODE sampler.

\section{Proof of the Proposition}
\textbf{Part (i): velocity form of the tilt.}
Define the per-noise-level tilted family $q_t(z)\propto p_{l,t}(z)\,\big[p^{\mathrm{RL}}_{s,t}(z)/p_{s,t}(z)\big]^{w}$. Its score is
\begin{equation}
\nabla_z\log q_t=s_l+w\,\big(s^{\mathrm{RL}}_s-s_s\big).
\label{eq:qscore_app}
\end{equation}
By Eq.~\eqref{eq:sv_app}, $v$ is affine in $s$ with model-independent coefficients, so score differences and velocity differences are proportional:
\begin{equation}
s^{\mathrm{RL}}_s-s_s=-\frac{1-t}{t}\big(v^{\mathrm{RL}}_s-v_s\big),
\end{equation}
and the velocity displacement induced by a score displacement $\Delta s$ is $-\frac{t}{1-t}\Delta s$. Composing the two factors,
\begin{equation}
-\frac{t}{1-t}\cdot w\,\big(s^{\mathrm{RL}}_s-s_s\big)
= w\,\big(v^{\mathrm{RL}}_s-v_s\big)
= w\,\Delta v_s,
\end{equation}
with the $t$-dependent factors cancelling exactly. Hence the velocity of $q_t$ is $v_l+w\,\Delta v_s$, the guided velocity of the main text. 

\textbf{Part (ii): exactness of the discrete step.}
Let the three models take the sampler step Eq.~\eqref{eq:sde_app} from the same state $z_t$. Each induces a Gaussian transition kernel $\mathcal N\big(\mu_\theta,\sigma_t^2I\big)$ with shared covariance and mean $\mu_\theta=a_tz_t+b_t v_\theta$ linear in the velocity. The pair's one-step log-ratio:
\begin{equation}
\Delta\ell(z')=\log\frac{\mathcal N(z';\mu^{\mathrm{RL}}_s,\sigma_t^2I)}{\mathcal N(z';\mu_s,\sigma_t^2I)},
\end{equation}
is affine in $z'$, because the quadratic terms cancel under the shared covariance. Multiplying a Gaussian density by an exponentiated affine function shifts its mean and leaves the covariance unchanged:
\begin{equation}
\mathcal N(\mu_l,\sigma_t^2I)\,e^{w\Delta\ell}\;\propto\;\mathcal N\big(\mu_l+w(\mu^{\mathrm{RL}}_s-\mu_s),\,\sigma_t^2I\big).
\label{eq:exactstep_app}
\end{equation}
Since $\mu^{\mathrm{RL}}_s-\mu_s=b_t\,\Delta v_s$ and $\mu$ is linear in $v$, the shifted mean equals $\mu(z_t;\,v_l+w\Delta v_s)$: substituting the guided velocity into the unmodified step realizes the reward-tilted one-step kernel exactly, with no linearization; the effective time weighting $b_t$ is inherited from the scheduler rather than designed. 

\section{The Pair Ratio Is the Source Value Function}
The exact step-wise object for sampling the tilted target is the soft value function $h_t(z)=\mathbb E\big[e^{r(z_0)/\beta}\,\big|\,z_t=z\big]$~\cite{uehara2025inference}. Under the shared forward kernel, the ratio of the pair's noisy marginals is exactly this object for the source alignment pair's own process: writing $p(z_t|z_0)$ for the shared kernel,
\begin{equation}
\begin{aligned}
\frac{p^{\mathrm{RL}}_{s,t}(z)}{p_{s,t}(z)}
&=\frac{\int p^{\mathrm{RL}}_s(z_0)\,p(z_t{=}z|z_0)\,dz_0}{p_{s,t}(z)}\\
&=\frac{1}{Z}\,\mathbb E_{p_s}\!\left[e^{r(z_0)/\beta}\,\middle|\,z_t=z\right],
\end{aligned}
\label{eq:valuefn_app}
\end{equation}
using $p^{\mathrm{RL}}_s(z_0)=p_s(z_0)e^{r(z_0)/\beta}/Z$. The RL run therefore amortized the value computation across all noise levels: the pair supplies, for free, the step-wise supervision that reward-based test-time methods must synthesize per step or train per model.

\section{Anchor Symmetry and Support}
For $w=1$ the target factorizes two ways:
\begin{equation}
p_l\cdot\frac{p^{\mathrm{RL}}_s}{p_s}\;=\;p^{\mathrm{RL}}_s\cdot\frac{p_l}{p_s}.
\label{eq:symmetry_app}
\end{equation} 
Reading (a): the host base model's prior, tilted by the source alignment pair's implicit reward. Reading (b): the source aligned model, tilted by the capability gap $p_l/p_s$ between large and small. Reading (b) makes the method weak-to-strong and fixes the support condition: a density ratio $p_A/p_B$ is readable only on $\mathrm{supp}(p_A)\cap\mathrm{supp}(p_B)$, and reading (b) assumes the benign inclusion $\mathrm{supp}(p_l)\supseteq\mathrm{supp}(p^{\mathrm{RL}}_s)$ (the strong model covers the weak one). Where the sampling trajectory leaves the shared support, the ratio degenerates to unconstrained network extrapolation; this is the mechanism behind the transfer boundary measured in the experiments, and it also explains why source-model weakness is harmless when transfer succeeds: on the overlap, the pair evaluates states the host base model visits in-distribution.

\section{Implementation Details}

We implement \method{} exactly as in Algorithm~\ref{alg:aligngraft}: at each step the host base model and the source alignment pair are read at the shared state, each through its own trained classifier-free guidance scale, and the alignment delta $w\,\Delta v_s$ is added to the host velocity before the sampler step. Throughout we use the deterministic ODE sampler, \ie, the variance coefficient of the Appendix~B step is set to zero ($\sigma_t{=}0$).

Table~\ref{tab:impl} lists the per-backbone sampling settings. For text-to-image (T2I) generation, the SD 3.5 backbones sample $40$ steps at $1024{\times}1024$ with CFG $4.5$, and the FLUX backbones sample $28$ steps at $1024{\times}1024$ with CFG $3.5$. For text-to-video (T2V) generation, the Wan-Fun backbones sample $30$ steps at $832{\times}480$ with $33$ frames per clip and CFG $6.0$. All SD 3.5 runs use one NVIDIA H100 GPU; the FLUX and Wan runs use one B300 GPU.

\begin{table}[t]
\centering
\small
\setlength{\tabcolsep}{4pt}
\begin{tabular}{@{}llccc@{}}
\toprule
Backbone (source\,$\to$\,host) & Mod. & Resolution & Steps & CFG \\
\midrule
SD 3.5 (M\,$\to$\,L)          & T2I & $1024{\times}1024$      & 40 & 4.5 \\
FLUX (dev\,$\to$\,Krea)      & T2I & $1024{\times}1024$      & 28 & 3.5 \\
Wan-Fun (1.3B\,$\to$\,14B)   & T2V & $832{\times}480$, 33\,f & 30 & 6.0 \\
\bottomrule
\end{tabular}
\caption{Per-backbone sampling configuration. Steps is the number of sampler discretization steps $N$ (Appendix~B); CFG is the classifier-free guidance scale, shared by the source alignment pair and the host within each family; ``33\,f'' denotes $33$ frames per clip.}
\label{tab:impl}
\end{table}

\section{Additional Experimental Details and Results}

This section provides further visual results, following the order of the main-text experiments.

\begin{figure}[t]
  \centering
  \includegraphics[width=0.7\linewidth]{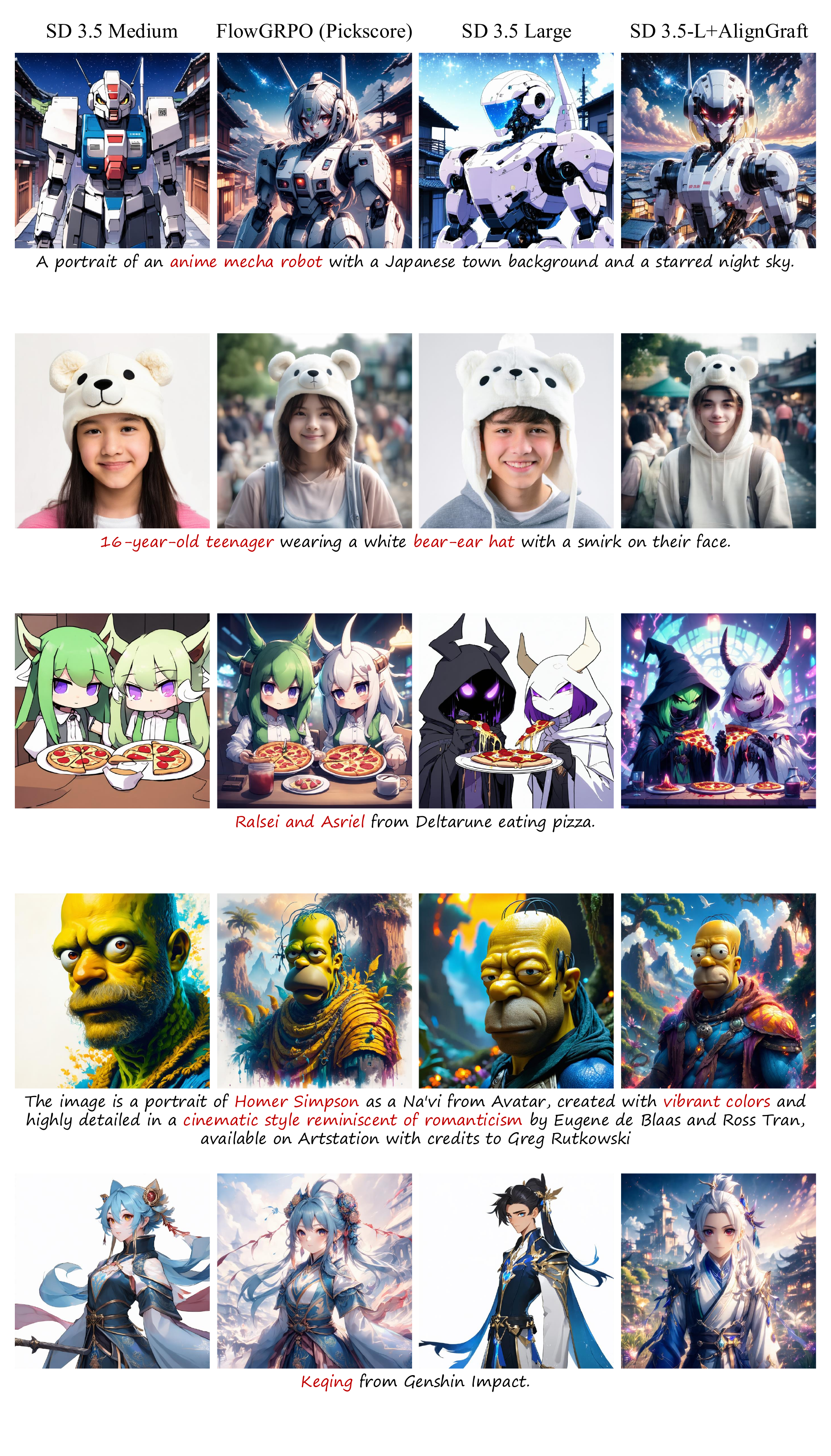}
  \caption{Additional PickScore transfers on SD 3.5 ($w{=}3$). Columns, left to right: source base (SD 3.5-Medium), source aligned (Flow-GRPO, PickScore), host base (SD 3.5-Large), and guided host (SD 3.5-Large\,+\,\method{}); each row is one prompt. The transferred samples inherit the preference reward (richer lighting, texture, and subject appeal) while retaining the host base model's native fidelity and layout.}
\label{fig:teacher_pick_1}
\end{figure}

\begin{figure}[t]
  \centering
  \includegraphics[width=0.7\linewidth]{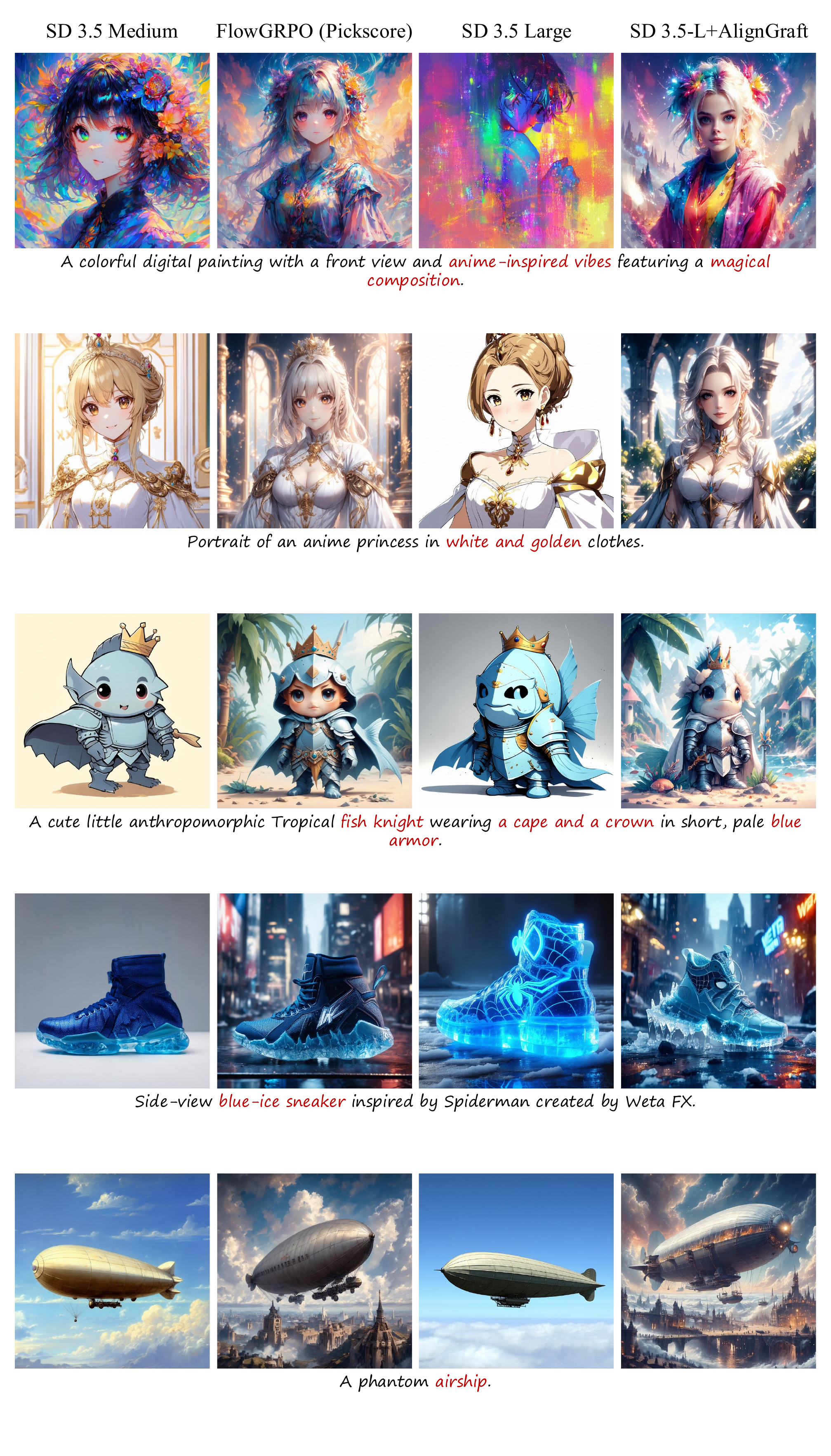}
      \caption{More PickScore transfers on SD 3.5 ($w{=}3$), continued from Figure~\ref{fig:teacher_pick_1}; same column layout (source base, source aligned, host base, and guided host).}
\label{fig:teacher_pick_2}
\end{figure}

\begin{figure}[t]
  \centering
  \includegraphics[width=0.7\linewidth]{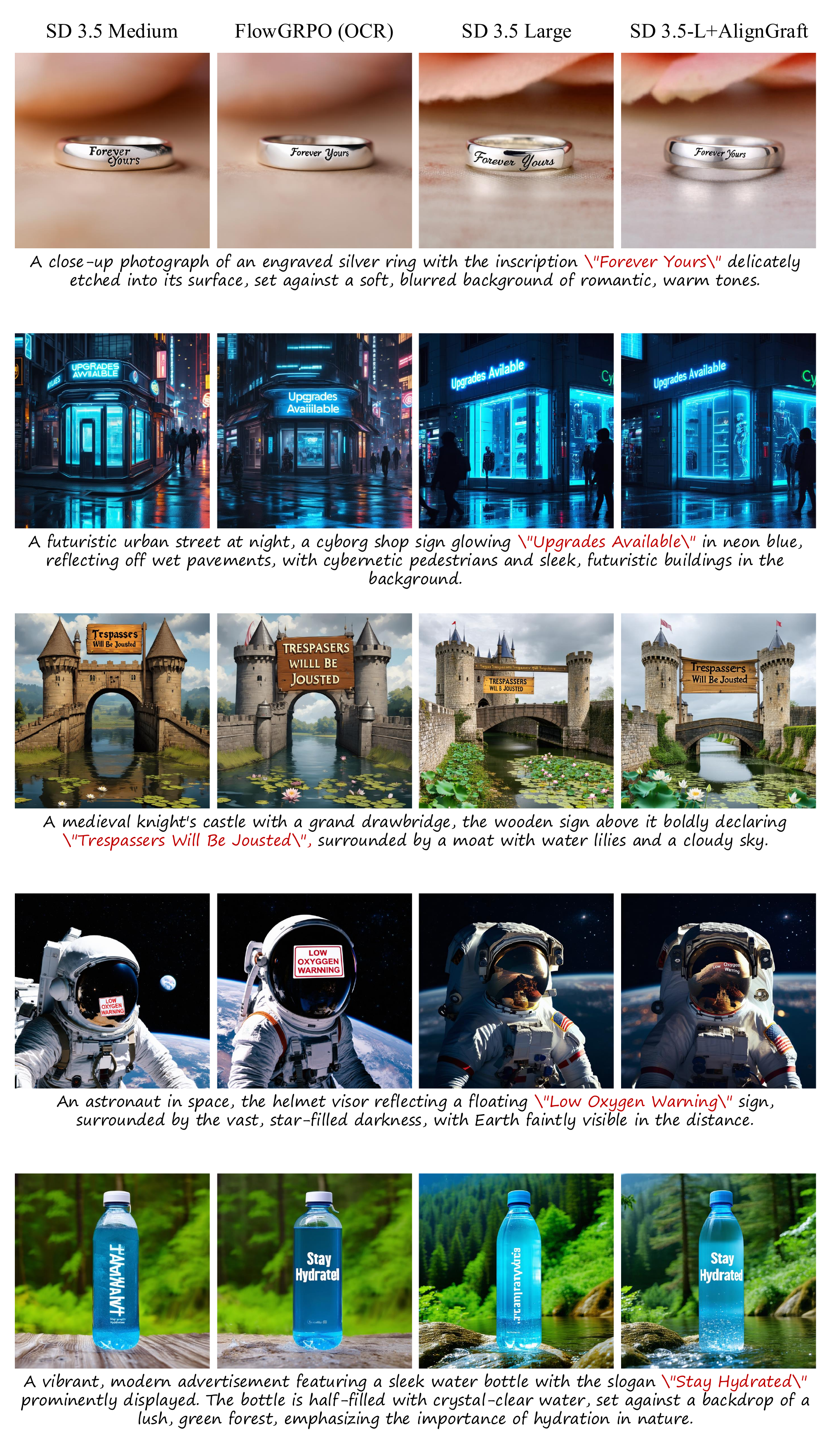}
      \caption{Additional OCR transfers on SD 3.5 ($w{=}1$). Columns, left to right: source base (SD 3.5-Medium), source aligned (Flow-GRPO, OCR), host base (SD 3.5-Large), and guided host (SD 3.5-Large\,+\,\method{}); each row is one prompt with its intended in-image text highlighted. The host base model renders malformed or mirrored glyphs, whereas the transfer produces legible in-image text matching the prompt while keeping the host base model's fidelity.}
\label{fig:teacher_ocr_1}
\end{figure}

\begin{figure}[t]
  \centering
  \includegraphics[width=0.7\linewidth]{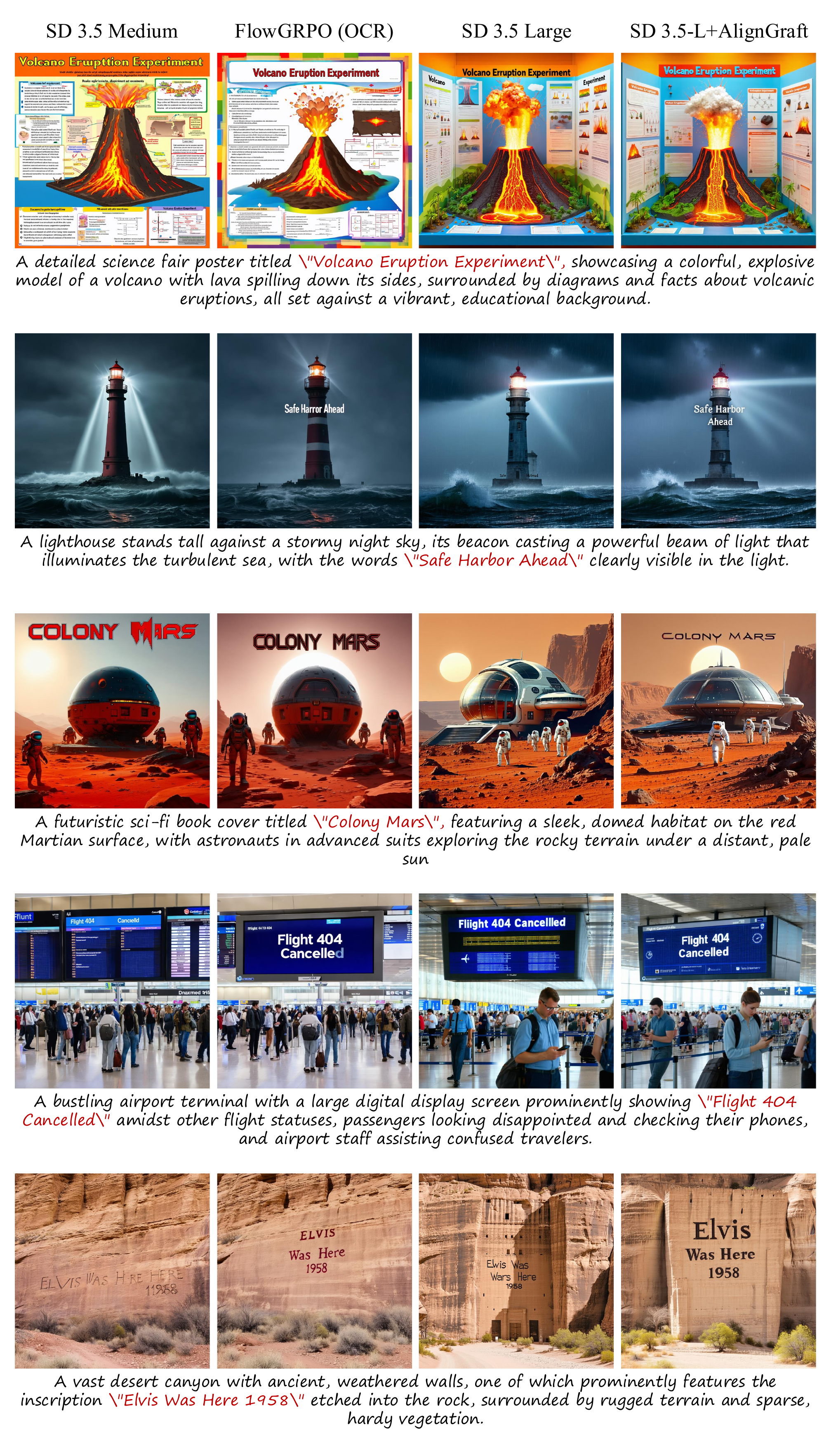}
      \caption{More OCR transfers on SD 3.5 ($w{=}1$), continued from Figure~\ref{fig:teacher_ocr_1}; same column layout (source base, source aligned, host base, and guided host).}
\label{fig:teacher_ocr_2}
\end{figure}

\begin{figure}[t]
  \centering
  \includegraphics[width=0.7\linewidth]{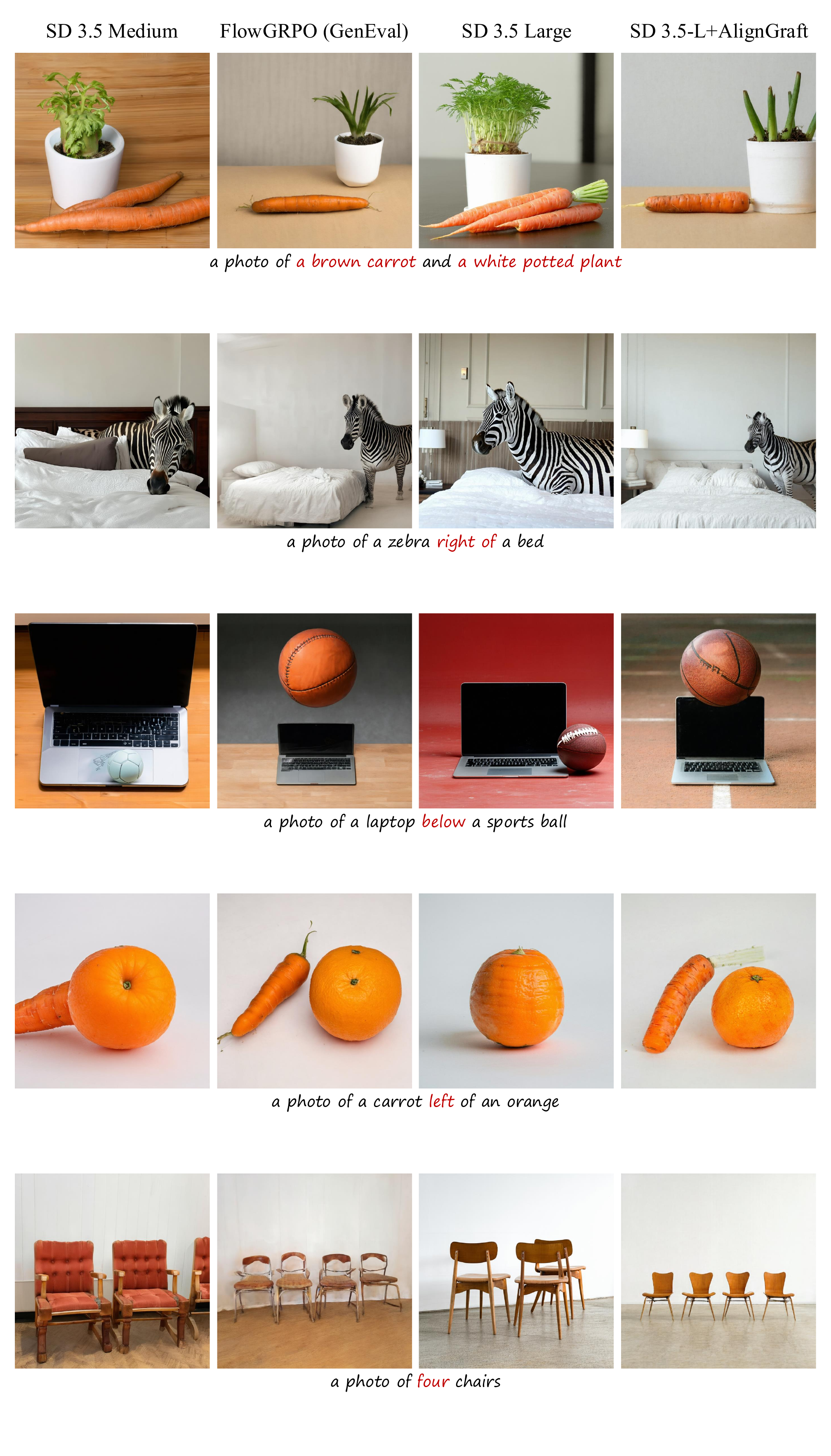}
      \caption{Additional GenEval transfers on SD 3.5 ($w{=}1$). Columns, left to right: source base (SD 3.5-Medium), source aligned (Flow-GRPO, GenEval), host base (SD 3.5-Large), and guided host (SD 3.5-Large\,+\,\method{}); each row is one prompt testing counting, spatial relation, or color binding. The transfer corrects compositional errors in object count and left/right/above/below arrangement that the host base model otherwise misses.}
\label{fig:teacher_geneval_1}
\end{figure}

\begin{figure}[t]
  \centering
  \includegraphics[width=0.7\linewidth]{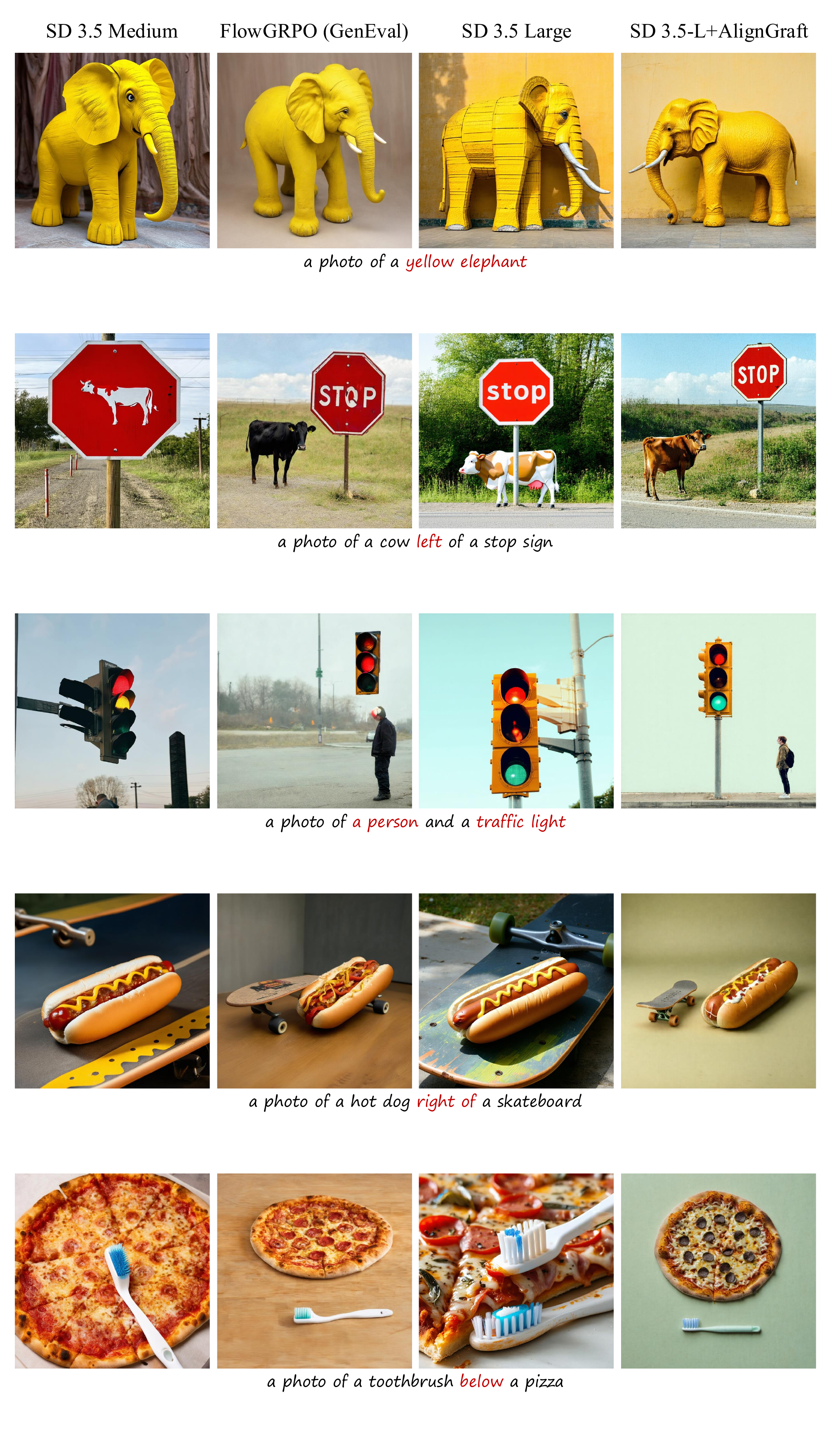}
      \caption{More GenEval transfers on SD 3.5 ($w{=}1$), continued from Figure~\ref{fig:teacher_geneval_1}; same column layout (source base, source aligned, host base, and guided host).}
\label{fig:teacher_geneval_2}
\end{figure}

\noindent\textbf{More qualitative results across reward sources.}\enspace
We give additional SD 3.5 transfers for each reward source, in the four-column layout of Figure~\ref{fig:teachers} (source base, source aligned, host base, and guided host). The PickScore source (Figures~\ref{fig:teacher_pick_1} and~\ref{fig:teacher_pick_2}) adds preference-aligned lighting, texture, and subject appeal; the GenEval source (Figures~\ref{fig:teacher_geneval_1} and~\ref{fig:teacher_geneval_2}) corrects compositional structure such as object count and left/right/above/below arrangement; and the OCR source (Figures~\ref{fig:teacher_ocr_1} and~\ref{fig:teacher_ocr_2}) renders legible in-image text where the host base model produces malformed glyphs. Across all three, the transfer moves each sample toward its source reward while the host keeps its native fidelity and resolution.

\begin{figure}[t]
  \centering
    \includegraphics[width=0.7\linewidth]{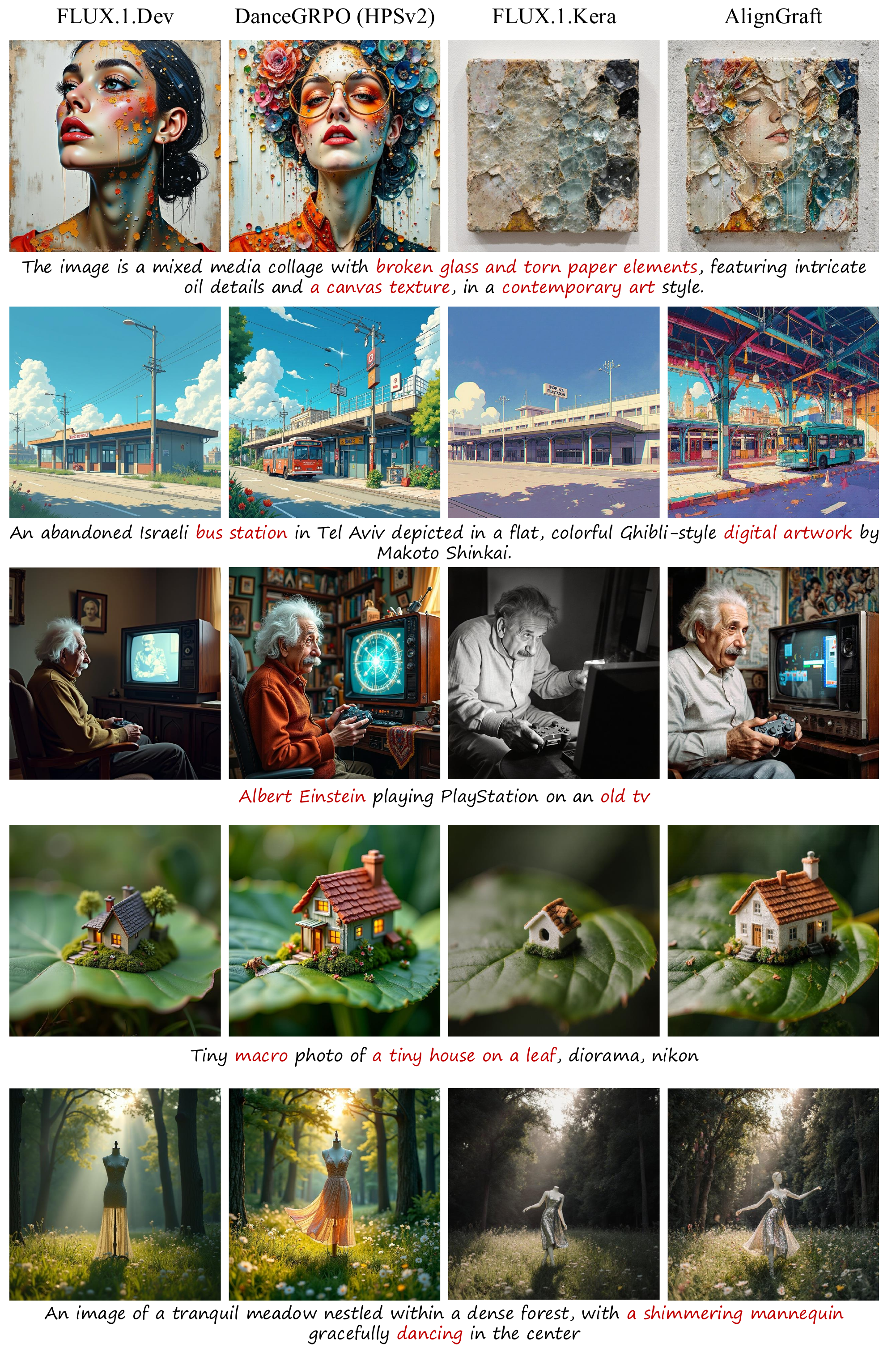}
  \caption{Additional cross-sibling transfers within the FLUX family ($w{=}1$). Columns, left to right: source base (FLUX.1-dev), source aligned (DanceGRPO, HPS v2.1), host base (FLUX.1-Krea), and guided host (FLUX.1-Krea\,+\,\method{}); each row is one prompt.}
\label{fig:flux_appendix}
\end{figure}

\begin{figure}[t]
  \centering
    \includegraphics[width=0.7\linewidth]{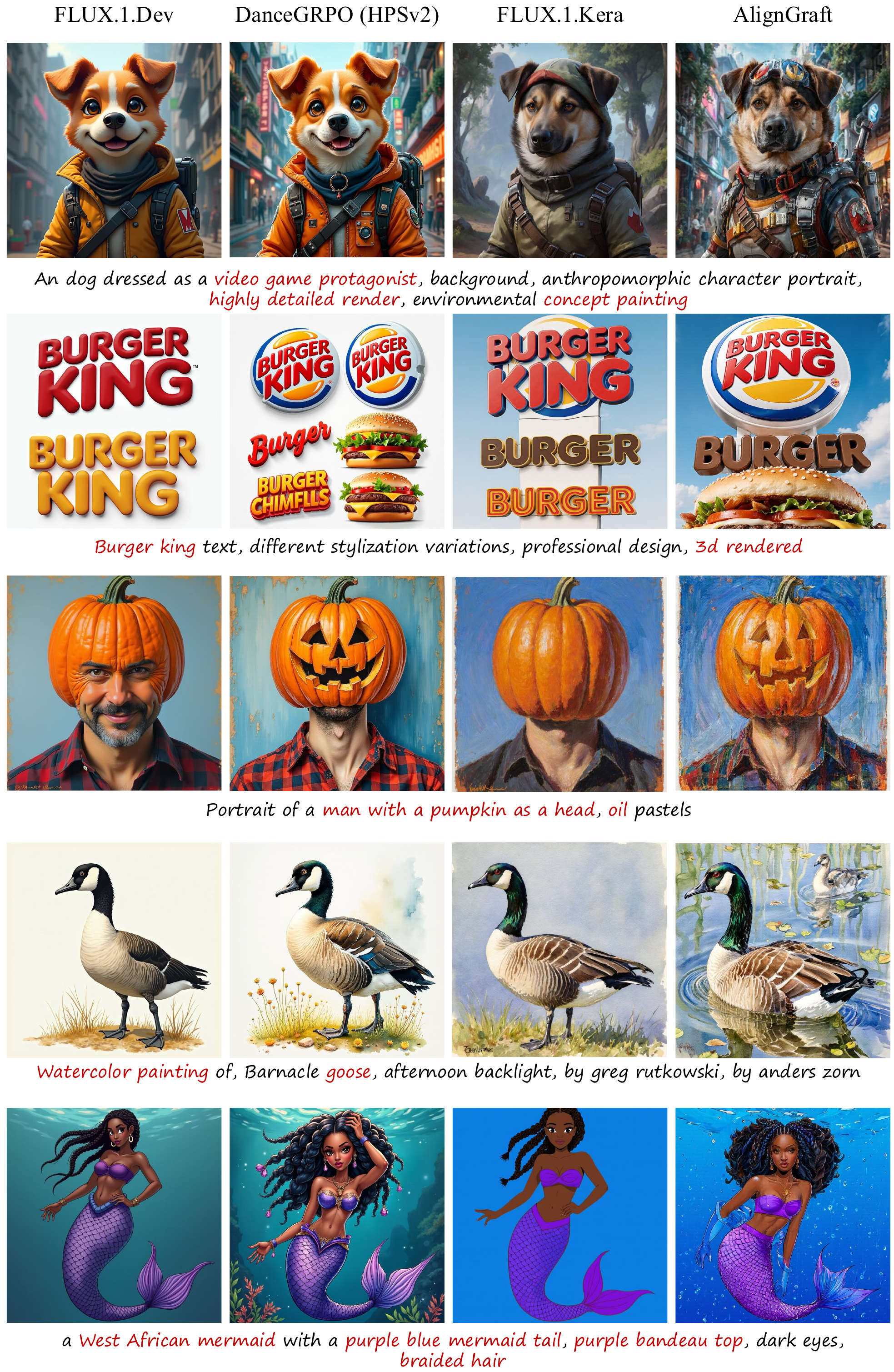}
  \caption{Additional cross-sibling transfers within the FLUX family ($w{=}1$). Columns, left to right: source base (FLUX.1-dev), source aligned (DanceGRPO, HPS v2.1), host base (FLUX.1-Krea), and guided host (FLUX.1-Krea\,+\,\method{}); each row is one prompt.}
\label{fig:flux_appendix2}
\end{figure}

\noindent\textbf{Cross-sibling transfer within the FLUX family.}\enspace
Beyond scaling up within a single lineage, the alignment delta transfers across independently post-trained siblings. A DanceGRPO pair on FLUX.1-dev guides the frozen, equally sized FLUX.1-Krea host base model, which shares the latent space but was never tuned for the reward. Figure~\ref{fig:flux_appendix} and Figure~\ref{fig:flux_appendix2} provide additional FLUX examples, complementing Figure~\ref{fig:flux} in the main text: at $w{=}1$ the transfer improves preference and aesthetic quality while preserving Krea's native style.

\begin{figure}[t]
  \centering
  \includegraphics[width=0.9\linewidth]{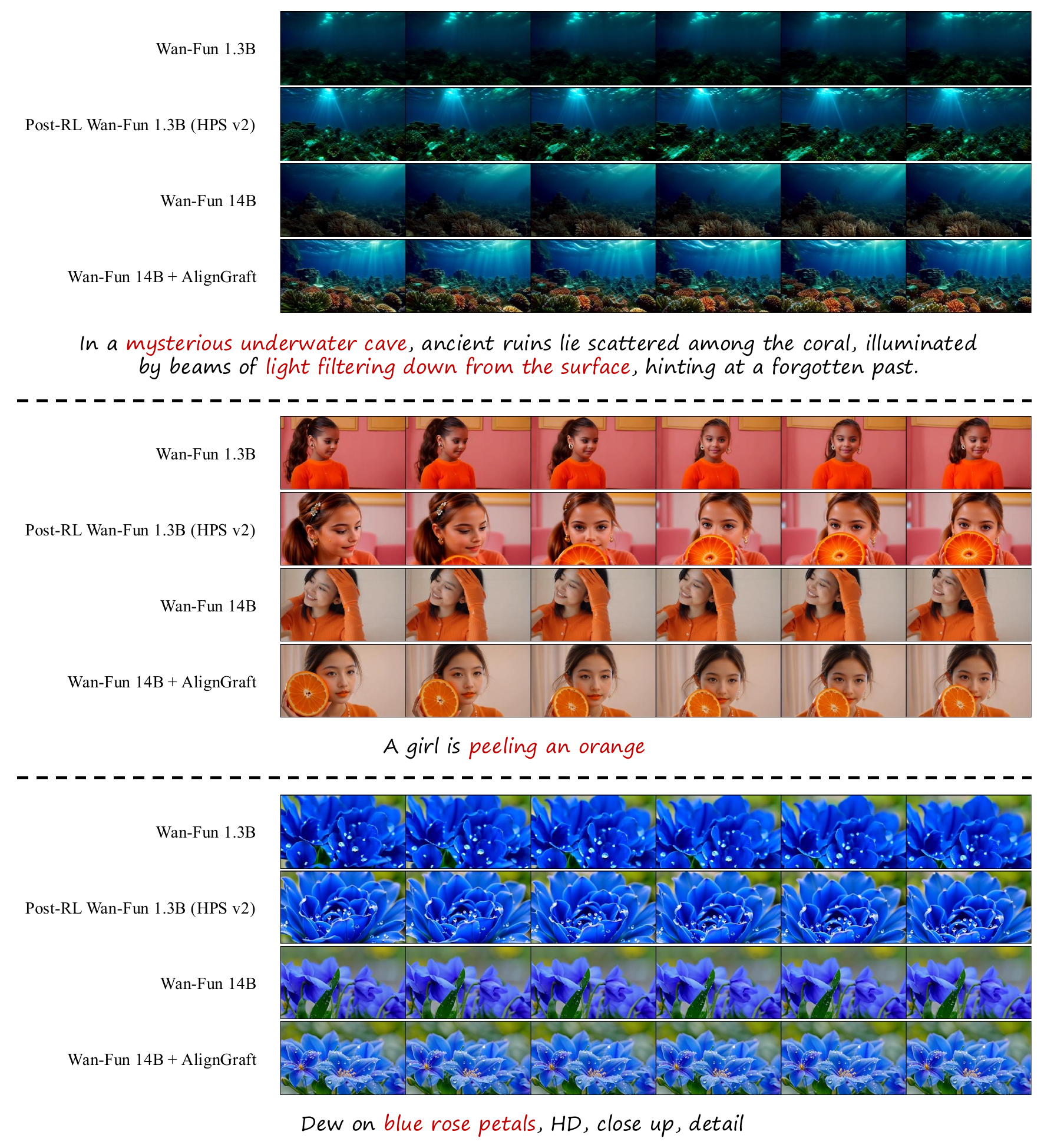}
      \caption{Additional text-to-video transfers on Wan-Fun with the HPS v2.1 source ($w{=}1$). Rows, top to bottom: source base (Wan-Fun-1.3B), source aligned (Wan-Fun-1.3B, reward backprop, HPS v2.1), host base (Wan-Fun-14B), and guided host (Wan-Fun-14B\,+\,\method{}); each row shows six frames of one clip, and blocks are separated by prompt. The transfer improves per-frame appeal and prompt adherence while preserving the host base model's motion.} 
\label{fig:wan_hps_1}
\end{figure}

\begin{figure}[t]
  \centering
  \includegraphics[width=0.9\linewidth]{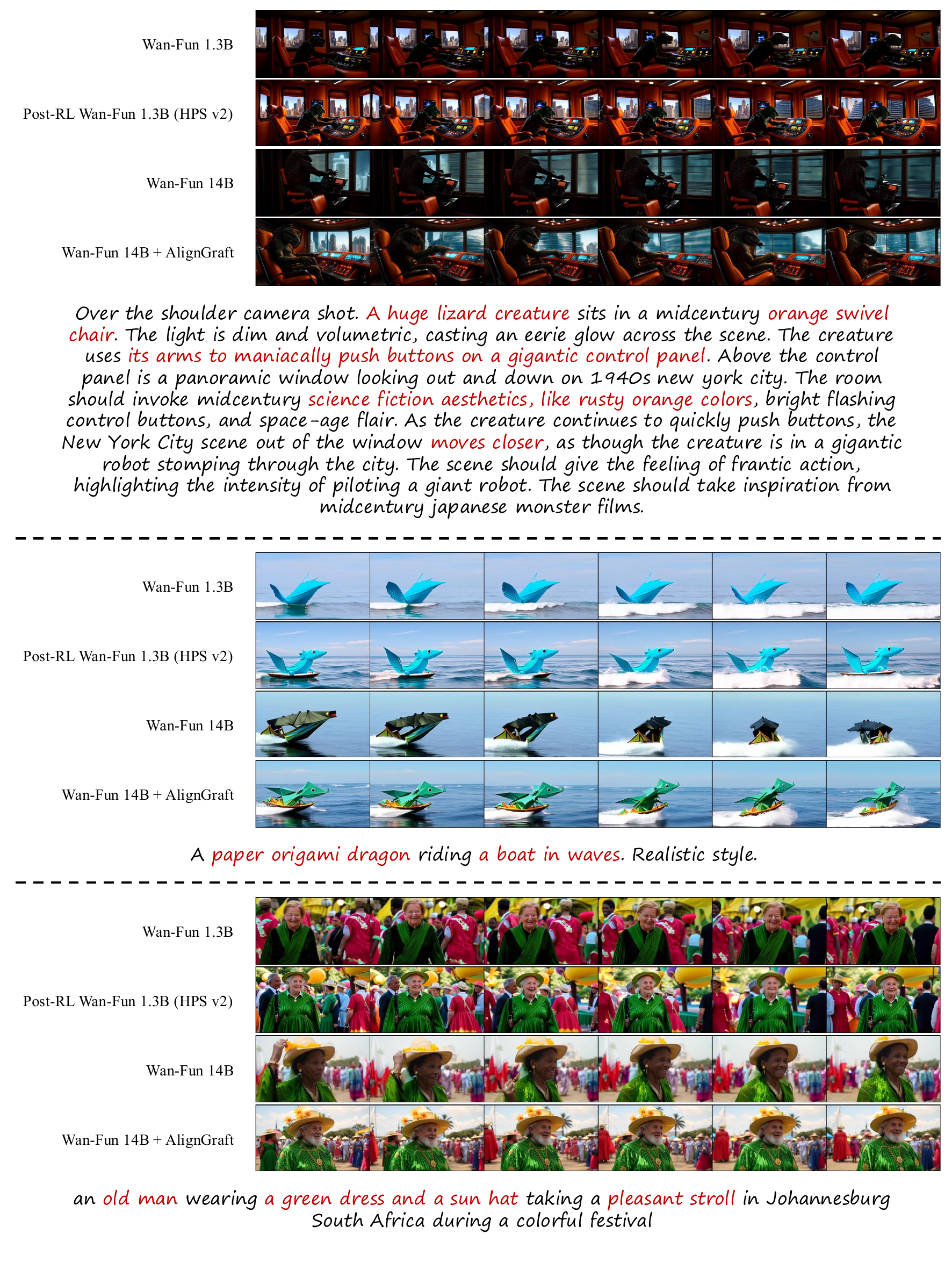}
      \caption{More HPS v2.1 text-to-video transfers on Wan-Fun ($w{=}1$), continued from Figure~\ref{fig:wan_hps_1}; same row layout (source base, source aligned, host base, and guided host), six frames per clip.}
\label{fig:wan_hps_2}
\end{figure}

\begin{figure}[t]
  \centering
  \includegraphics[width=0.9\linewidth]{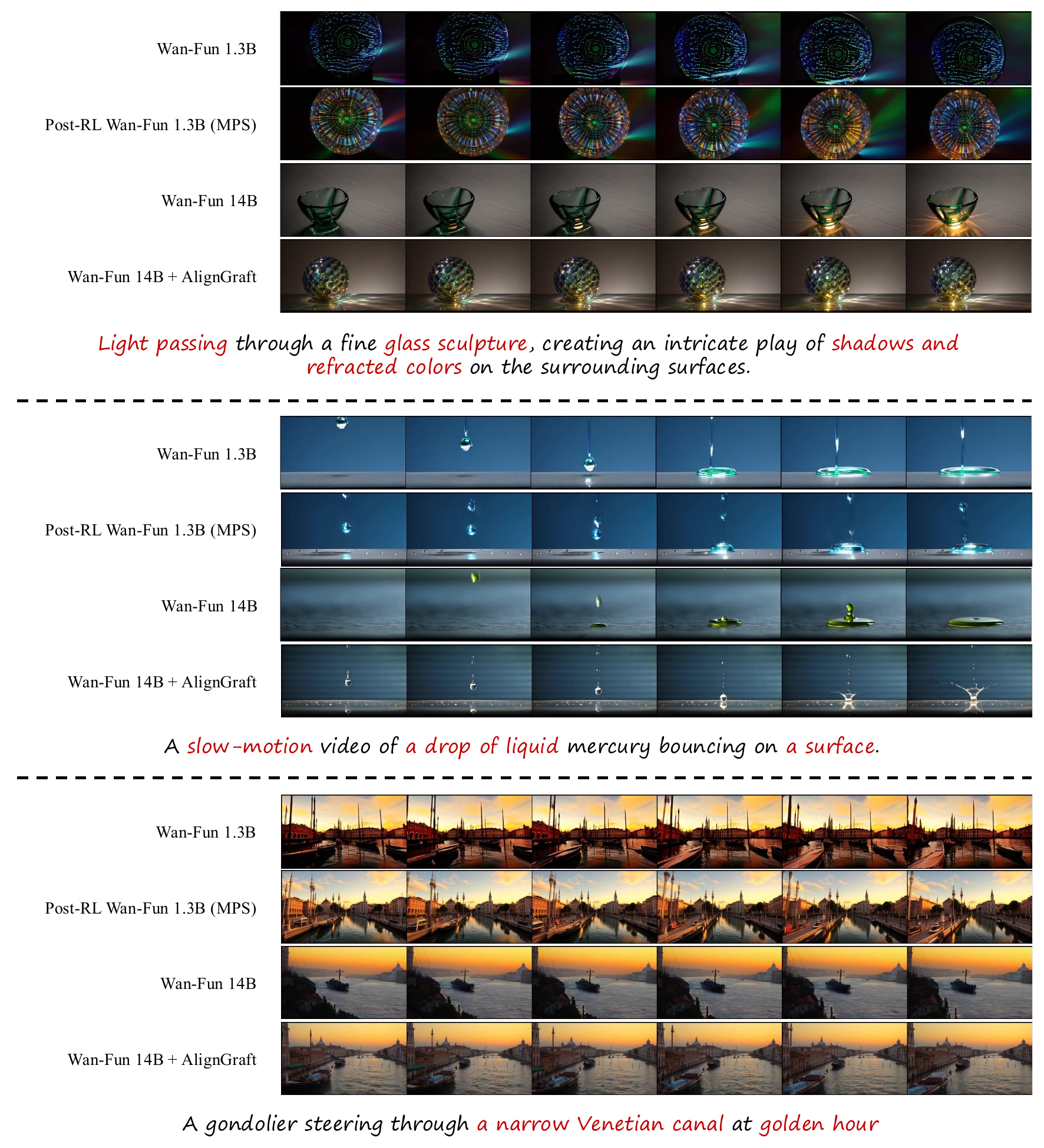}
      \caption{Additional text-to-video transfers on Wan-Fun with the MPS source ($w{=}1$). Rows, top to bottom: source base (Wan-Fun-1.3B), source aligned (Wan-Fun-1.3B, reward backprop, MPS), host base (Wan-Fun-14B), and guided host (Wan-Fun-14B\,+\,\method{}); each row shows six frames of one clip, and blocks are separated by prompt. The transfer improves motion coherence and prompt adherence.}
\label{fig:wan_mps_1}
\end{figure}

\noindent\textbf{Video-domain transfer on Wan.}\enspace
The same sampling rule extends unchanged to text-to-video. Figures~\ref{fig:wan_hps_1} and~\ref{fig:wan_hps_2} show additional Wan-Fun clips under the HPS v2.1 source, and Figure~\ref{fig:wan_mps_1} under the MPS source: guiding the frozen Wan-Fun-14B host base model improves per-frame appeal and prompt adherence across the six sampled frames while preserving the motion of the underlying clip, consistent with the quantitative results in the main text.

\begin{figure}[t]
  \centering
  \includegraphics[width=0.9\linewidth]{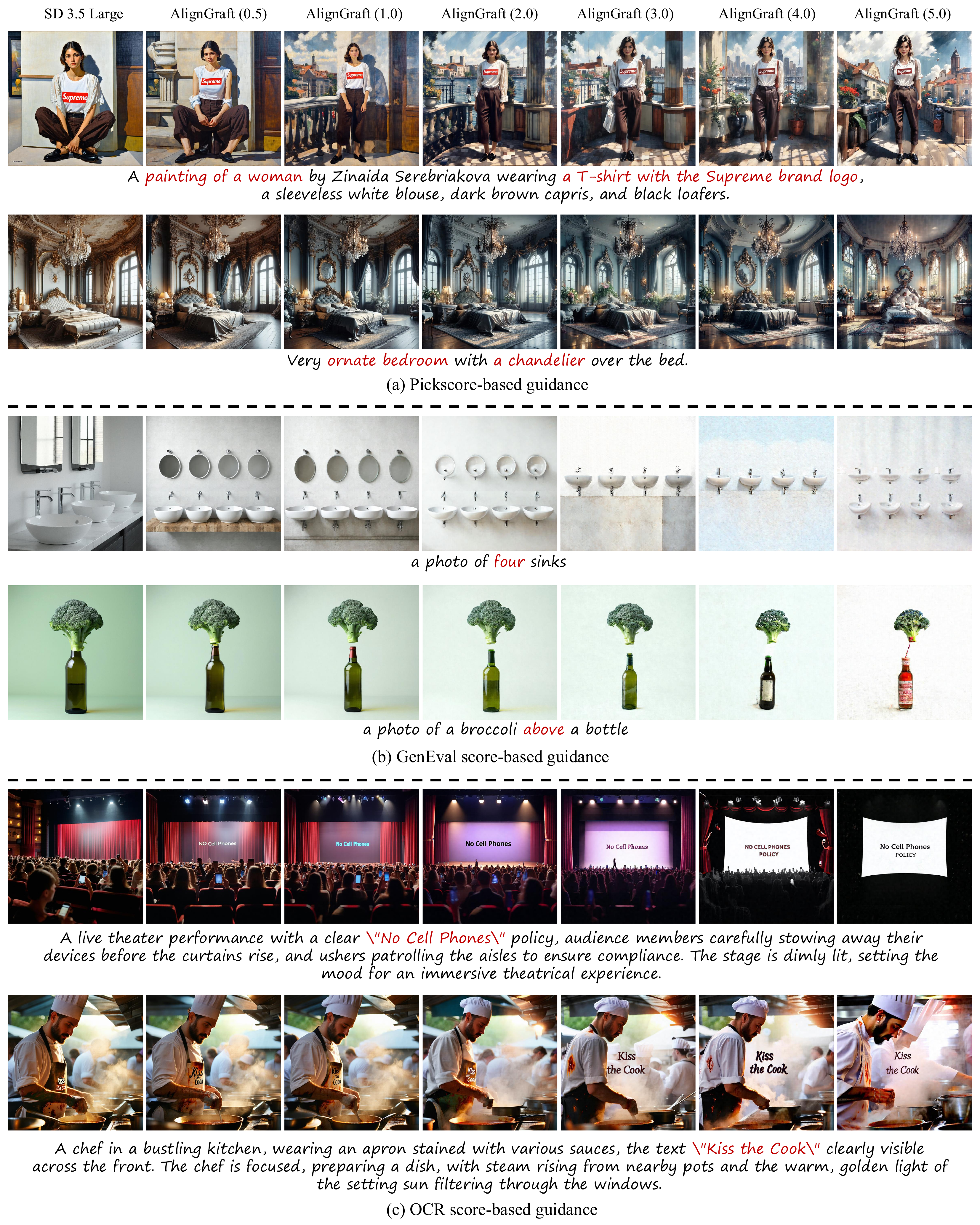}
      \caption{Effect of the guidance strength $w$ on SD 3.5. The leftmost column is the frozen SD 3.5-Large model ($w{=}0$); the remaining columns apply \method{} at $w\in\{0.5,1,2,3,4,5\}$. Blocks: (a) PickScore, (b) GenEval, and (c) OCR reward sources. Reward alignment strengthens with $w$ up to a broad peak and then degrades into over-optimization at large $w$ (saturated color, collapsed layout, distorted text), the qualitative counterpart of the inverted-U in Figure~\ref{fig:search}.}
\label{fig:guidance_varies}
\end{figure}

\noindent\textbf{Effect of guidance strength.}\enspace 
Figure~\ref{fig:guidance_varies} visualizes sweeping the strength $w$ on SD 3.5 for the three reward sources, from the host base model ($w{=}0$) up to $w{=}5$. At moderate $w$ the transferred reward and the visual quality improve together. Extrapolation helps only up to a point: larger $w$ induces reward hacking, where the measured reward keeps rising yet image quality degrades, because the sample collapses onto the reward model's idiosyncratic bias rather than genuine quality. This bias is source-specific and evident in Figure~\ref{fig:guidance_varies}; for instance, the PickScore transfer over-saturates into gaudy, high-contrast colors, while the GenEval transfer flattens the scene onto a plain white background, each a shortcut its own scorer over-rewards. These two opposing trends, reward rising monotonically while perceptual quality first improves and then breaks down, are precisely why quality traces an inverted-U in $w$: it peaks while the gain is still genuine and falls once reward hacking dominates, the qualitative counterpart of the inverted-U measured in the main text (Figure~\ref{fig:search}).

\section{Social Impact}

\method{} makes reward alignment portable: one cheap alignment run on a small model can steer a larger, frozen model at test time, lowering the cost of alignment and broadening its access without modifying the host base model's weights. As with other controllable generation methods, a few potential concerns are worth noting. Because the method is agnostic to the objective it transfers, it could in principle be used to steer a model toward less desirable content, and its extension to video somewhat broadens the settings where this might arise; strong guidance may also emphasize the particular biases of the source reward. We therefore encourage responsible use and discourage applying \method{} to produce harmful, misleading, or non-consensual content, together with transferring rewards whose objectives and data are reasonably vetted, pairing generation with provenance, watermarking, and moderation, and keeping the guidance strength within the anchored regime where alignment improves quality rather than over-optimizing the reward. While the underlying generative backbones typically include a built-in safety checker, such filters are not perfect, so we recommend using \method{} together with these safeguards.

\section{Limitations}
\method{} depends on the full \emph{source alignment pair} $(\theta_s,\theta_s^{\mathrm{RL}})$, since the alignment signal is their difference $\Delta v_s=v_s^{\mathrm{RL}}-v_s$ rather than the source aligned model alone. This dependence is easy to satisfy: the base is usually public, and the source aligned model is either released with it or, being small, cheap to re-train; it becomes limiting only when a post-RL checkpoint lacks a matching base.

\end{document}